\documentclass[11pt]{article}
\usepackage[letterpaper,margin=1in]{geometry}
\usepackage[final]{changes}

\PassOptionsToPackage{numbers,compress}{natbib}

\usepackage{natbib}

\usepackage{amsmath,amssymb,amsfonts}
\usepackage{amsthm}

\usepackage{algorithm}
\usepackage{algpseudocode}

\usepackage{booktabs}
\usepackage{hyperref}
\usepackage{url}
\usepackage{nicefrac}
\usepackage{microtype}
\usepackage{xcolor}

\makeatletter
\newcommand{\listappendicesname}{List of Appendices}
\newcommand{\listofappendices}{%
  \section*{\listappendicesname}
  \addcontentsline{toc}{section}{\listappendicesname}
  \@starttoc{apc}%
}
\makeatother

\newtheorem{definition}{Definition}
\newtheorem{assumption}{Assumption}
\newtheorem{proposition}{Proposition}
\newtheorem{lemma}{Lemma}
\newtheorem{theorem}{Theorem}
\newtheorem{corollary}{Corollary}
\newtheorem{remark}{Remark}
\newtheorem{example}{Example}

\usepackage{graphicx}
\usepackage{subcaption}

\title{Trust-Region Optimization for Performative Prediction without Response Modeling}

\author{%
  Siqi~Du\thanks{Siqi Du and Roy Dong are with the Department of
  Industrial and Enterprise Systems Engineering, University of Illinois
  Urbana-Champaign, Urbana, IL, USA.
  Email: \texttt{\{siqidu3,roydong\}@illinois.edu}.}
  \and
  Miaolan~Xie\thanks{Miaolan Xie is with the Edwardson School of
  Industrial Engineering, Purdue University, West Lafayette, IN, USA.
  Email: \texttt{miaolanx@purdue.edu}.}
  \and
  Roy~Dong\footnotemark[1]%
}
\date{}

\begin{document}

\maketitle

\begin{abstract}
Performative prediction studies learning problems in which deployed decisions change the distribution of future data. Optimizing under this feedback requires understanding not only how a decision affects the loss, but also how the population responds to that decision. Existing approaches often make this problem tractable by imposing structure on the population response, but this can limit applicability when responses are poorly captured by a fixed model class. Generic black-box optimization avoids response modeling, but discards structure that is readily available in this setting: the loss function itself is often known in performative prediction.

We introduce Performative Trust-Region Optimization (PTRO), which bridges these two approaches by optimizing performative risk without specifying or fitting a response model while still exploiting the known loss. The key idea is sampling from the current population to evaluate the known loss at nearby decisions, and use additional deployments only to estimate the effect of population response. This gives a local approximation of performative risk without committing to a model of how the population responds. PTRO then uses a trust-region mechanism to determine and adapt how far this local information can be trusted.

Under moment-bounded losses and suitable smoothness and estimator-accuracy conditions, we establish high-probability convergence to an $\epsilon$-stationary point of the performative risk in $O(\epsilon^{-2})$ iterations, with $O(d^{3}\epsilon^{-4})$ samples per iteration, where $d$ is the decision dimension and $q\geq2$ is the loss-moment order. 

Experiments illustrate methods built around a specified response model can be more sample-efficient when their structural assumptions are accurate, whereas PTRO remains effective as the population response is poorly captured by a fixed model class. Ablations further show that both exploiting
the known loss and adaptive trust-region updates are
important for controlling the accuracy and efficiency.



\end{abstract}

\section{Introduction}
Deploying a prediction model can change the population on which it is
evaluated. Unlike standard supervised learning, where the
data distribution is treated as fixed, in many real systems a
prediction affects decisions made about individuals, which can in turn
change their future behavior or outcomes. For example, a classifier that predicts crime recidivism influences the opportunities available to individuals~\cite{dressel2018accuracy}, a recommendation  system can influence what users consume and subsequently engage
with. A model that performs
well on the training distribution may perform poorly on the
distribution induced by its deployment. 
Performative
prediction~\citep{perdomo2020performative} formalizes this feedback through
a distribution map $D(\cdot)$: deploying a decision $x$
induces a data distribution $D(x)$. Define the \textit{performative risk} of a decision $x$ as: 
\begin{align*}
    PR(x) = \mathbb{E}_{z \sim D(x)} [\ell(z;x)]
\label{eq:PR}
\end{align*}
where $\ell$ is the loss function. Optimizing this objective requires
accounting for both the direct effect on loss and the indirect effect of the distribution shift that the decision induces.

A natural approach is to model how the population responds and use the
estimated response to optimize performative risk. Several existing methods
follow this approach by assuming a structured distribution family or a
parametric response mechanism and estimating its parameters from deployment
data~\citep{miller2021outside,izzo2021learn,lin2023plug,hamaguchi2026adaptive}. But this can be challenging for
heterogeneous populations: different groups may respond through different
mechanisms, producing a mixture whose components and response parameters
are unknown. Fitting an appropriate response model can
then become a challenging part of the optimization problem.

Motivated by the question:
\emph{Can we optimize performative risk without specifying or estimating
a model of the population response?}
We consider a setting in which the learner knows the loss function and
can collect samples from $D(x)$ by deploying $x$, but does not assume a
known parametric representation of $D(\cdot)$. Instead, we assume
statistical conditions on the induced loss and smoothness conditions on
the risk. This condition can accommodate unknown parametric models, unknown mixture structures, or response
distributions with no fixed parameterization, whenever the required loss moments
and regularity conditions hold. 
Finding the optimal decision directly is challenging. Even when the loss is strongly convex in the decision and the response map is smooth, the resulting performative risk can be nonconvex as shown in Example~\ref{exp:nonconvex}.  So we target a relaxation of it: an $\epsilon$-stationary point ($\epsilon$-SP) of the performative risk,
$$\|\nabla PR(x)\|\le\epsilon.$$ 
Define the
decoupled risk
    $R(x_1,x_2)
    := \mathbb{E}_{z \sim D(x_2)}[\ell(z;x_1)],$
where $x_1$ is the decision used to evaluate the loss and $x_2$ is the
decision that induces the population distribution. 
The key observation is that the loss function is often known and the population response is not. We use this structure in \emph{Performative Trust-Region Optimization}
(PTRO) : its local approximation evaluates the empirical loss on the current population and approximates the additional effect of the population response using finite differences under nearby deployments, without specifying or fitting a response model.

 At each iteration, PTRO approximately minimizes this model within a trust region. Then it evaluates the trial decision by comparing the empirical
risk reduction with the reduction predicted by the local approximation.
This agreement determines whether the step is accepted and how the
trust-region radius is updated.

As a result, we establish high-probability convergence of PTRO to an $\epsilon$-stationary point of the performative risk under the two representative settings.
In particular, PTRO requires $O(\epsilon^{-2})$ iterations. Under our estimator constructions, the
per-iteration sample complexity is
$O(d^{3}\epsilon^{-4})$ under moment-bounded risk.

Therefore, our main contributions are summarized as follows.
\begin{itemize}
\item We develop a performative optimization framework  without requiring population response to follow a prespecified model, while exploiting the known loss structure to avoid treating the entire objective as a black box.

\item We develop PTRO and establish convergence to stationary points, shifting the required assumptions from prior knowledge of the form of the population response to regularity conditions on the induced performative risk.

\item Experiments show  that structured response models can be more sample-efficient when well specified, while PTRO remains effective under poorly modeled responses. Ablations further demonstrate the benefits of exploiting the known loss and adapting the trust-region radius.
\end{itemize}

\section{Related work}
We discuss our results in the context of background and related work in Appendix~\ref{appdx:background}. In summary, existing methods for performative optimality~\citep{miller2021outside,izzo2021learn,lin2023plug,hamaguchi2026adaptive} typically assume that $D(\cdot)$ belongs to a known parametric or location family, and then estimate the corresponding parameters from deployment data. They then optimize the performative objective using gradient-based updates. In contrast, zeroth-order methods~\citep{liu2023two,hikima2025guided} do not require a parametric response model and estimate optimization directions from function evaluations. We summarize these works in Table~\ref{table:comparison_existing_method}.

The case of unknown $D(\cdot)$ is related to black-box optimization~\citep{flaxman2004online,ghadimi2013stochastic} and standard trust-region
methods~\citep{gratton2018complexity,blanchet2019convergence,cao2024first,jin2025sample,scheinberg2025stochastic}. A direct application of these methods would treat the entire
performative risk as an unknown objective, ignoring the fact that the loss
function $\ell$ is often known. PTRO preserves the adaptive validation mechanism of standard trust-region methods to avoid unreliable updates, and uses the known loss structure directly into the local approximation to better exploit the structure of performative prediction.

\begin{table}[t]
\caption{
Comparison of representative methods for performative optimization.\\
Here, $\epsilon$-PO denotes
$PR(x)-PR(x_{\mathrm{PO}})\le\epsilon$,
HP denotes a high-probability guarantee and Exp denotes a guarantee in
expectation.\\
$^{\ddagger}$Plug-in does not require convexity because,  it assumes
that the resulting plug-in optimization problem can be solved after fitting the response model.\\
$^{\S}$The displayed PerfGD rates correspond to its one-dimensional
Gaussian analysis.
}
\centering
\tiny
\begin{tabular}{lccccc}
\hline
Algorithm
& Target
& \shortstack{Convexity\\of $PR$}
& \shortstack{Parametric\\response model}
& \shortstack{Iteration\\complexity}
& \shortstack{Samples\\per iteration}
\\
\hline

PTRO (this work)
& $\epsilon$-SP (HP)
& Nonconvex
& NA
& $O(\epsilon^{-2})$
& $O(d^{3}\epsilon^{-4})$
\\

Two-stage~\citep{miller2021outside}
& $\epsilon$-PO (HP)
& Strongly convex
& \checkmark
& $O(d\epsilon^{-1})$
& $O(1)$
\\

PerfGD~\citep{izzo2021learn}
& $\epsilon$-SP
& Convex
& \checkmark
& $O(\epsilon^{-4})^{\S}$
& $\widetilde O(\epsilon^{-10})^{\S}$
\\
Plug-in~\citep{lin2023plug}
& $\epsilon$-PO (HP)
& Not required$^{\ddagger}$
& \checkmark
& $O(d_\beta \epsilon^{-2})$
& Depends on response-model fitting
\\

DFO$(\lambda)$~\citep{liu2023two}
& $\epsilon$-SP (Exp)
& Nonconvex
& NA
& $O(d^2\epsilon^{-6})$
& $O(\log T)$
\\

DSA-PGM~\citep{hamaguchi2026adaptive}
& $\epsilon$-SP (Exp)
& Nonconvex
& \checkmark
& $O(\epsilon^{-2})$
& $O(d^{5/2}\epsilon^{-3})$
\\
Guided-ZO~\citep{hikima2025guided} &$\epsilon$-SP (Exp)
& Nonconvex
& NA
& $O(d^4\epsilon^{-6})$
& $O(1)$
\\


\hline
\end{tabular}
\label{table:comparison_existing_method}
\end{table}

\section{Problem Setup}

\paragraph{Objective.}
We consider decisions $x \in \mathcal{X}\subset\mathbb{R}^d.$ Deploying $x$ induces
a data distribution $D(x)$, which leads to the performative risk
$
    PR(x) := \mathbb{E}_{z \sim D(x)}[\ell(z;x)],
$
where $\ell(z;x)$ is a known loss function.  Since $PR$
may be nonconvex, our goal is to find an $\epsilon$-stationary point ($\epsilon$-SP), which is a decision $x$ satisfying
$
    \|\nabla PR(x)\| \leq \epsilon.
$

\begin{assumption}[Regularity of the performative risk]
\label{assp:PR_smooth}
The performative risk is bounded below:
$
    PR_* := \inf_{x \in \mathcal{X}} PR(x) > -\infty.
$
Moreover, $PR$ is continuously differentiable and its gradient
is $L$-Lipschitz:
$
    \|\nabla PR(x)-\nabla PR(x')\|
    \leq L\|x-x'\|,
    \qquad \forall x,x' \in \mathcal{X}.
$
\end{assumption}


\paragraph{Deployment and sampling.}
The learner can deploy a decision $x$ and collect independent samples
from $D(x)$. We do not assume a known parametric representation of
$D(\cdot)$ or access to its density or derivatives. Given an observation
$z$, the learner can evaluate the loss $\ell(z;x')$ and its gradient
with respect to $x'$ at any candidate decision $x'$.

Recall the decoupled risk
$R(x_1,x_2)
    := \mathbb{E}_{z \sim D(x_2)}[\ell(z;x_1)],$
samples
collected under $x_2$ can be used to estimate $R(x_1,x_2)$
at different values of $x_1$ without additional deployments.

\paragraph{Problem scope: Moment-bounded losses.}
We study performative prediction problems in which the loss has a uniformly bounded moment across evaluation and deployment decisions.  
\begin{equation}
    \sup_{x_1,x_2\in\mathcal{X}}
    \mathbb{E}_{z\sim D(x_2)}
    \bigl[|\ell(z;x_1)|^q\bigr]
    \le L_q \qquad \text{for some }q \geq 2 \text{ and } L_q < \infty
\label{eq:moment_bdd_loss}   
\end{equation}
 This condition does not require $D(\cdot)$ to belong to a specified parametric family. 
It allows non-Gaussian responses, mixtures of distributions from different families, and heavy-tailed responses, provided that the induced loss satisfies the uniform moment bound. The condition holds immediately for uniformly bounded losses. It also holds for losses that grow at most polynomially in the data, uniformly over the decision domain, when the response distributions have sufficiently high uniformly bounded moments.

For example, suppose the population consists of heterogeneous response types, $D(x)=\int D_\omega(x)\,dP(\omega),$ where $\omega$ may index latent groups or even individual-specific response
mechanisms. If each component satisfies a uniform $q$-th loss-moment bound
$L_q(\omega)$ and
$\int L_q(\omega)\,dP(\omega)<\infty,$
then the induced population distribution also satisfies the moment condition
with
    $L_q=\int L_q(\omega)\,dP(\omega).$

Thus, PTRO does not require the learner to identify the latent response types,
estimate their proportions, or specify a parametric form for
$D_\omega(\cdot)$. 
Appendix~\ref{Existing_Response Models_as_Special_Cases} further shows that
several structured response models from prior work are covered by our
assumptions.

\section{Performative Trust-Region Optimization}
\label{sec:PTRO}
We introduce Performative Trust-Region Optimization (PTRO). At iteration $k$, the deployed decision $x_k$ induces the current population
$D(x_k)$, from which PTRO collects samples and constructs a local model of the
performative risk. Rather than treating the entire objective as a black box,
PTRO exploits the decoupled structure of performative prediction: the effect
of changing the decision on the known loss is evaluated directly on samples
from $D(x_k)$, and locally approximate the unknown distributional-response effect.
PTRO approximately minimizes this performative local approximation within a trust
region to obtain a trial step. Then it collects fresh samples under the trial
decision and compares the realized performative-risk reduction with the
predicted reduction. This comparison determines whether the step is accepted
and how the trust-region radius is updated. So the trust-region radius controls how far PTRO trusts the local
information collected under the current decision.

In this section, we first construct the decoupled local approximation,
then describe the trust-region updates and present the
algorithm.
\subsection{Local approximation}

Recall that $R(x_1,x_2)$ separates the decision used to evaluate the
loss from the decision that induces the population distribution. $PR(x_k+s) = R(x_k+s, x_k+s).$ 
Then, for a candidate step $s$ at the current decision $x_k$, adding and subtracting $R(x_k+s,x_k)$ gives
\begin{equation}
PR(x_k+s)
=
\underbrace{R(x_k+s,x_k)}_{\text{expected loss on the current population}}
\quad+
\underbrace{
R(x_k+s,x_k+s)-R(x_k+s,x_k)
}_{\text{effect of population response}}.
\label{eq:performative_decomp}
\end{equation}

The decomposition separates the part that can be evaluated using the
current population from the part that depends on how the population
changes under a new deployment.

For later use, we first define an empirical estimator of the
decoupled risk. Given a pair $(x_1,x_2)$, 
\begin{equation}
f_m(x_1,x_2;\xi)
:=
\frac{1}{m}
\sum_{j=1}^{m}
\ell(z_j;x_1),
\qquad
z_j \stackrel{\mathrm{i.i.d.}}{\sim} D(x_2),
\label{eq:empirical_decoupled_risk}
\end{equation}
Here, $\xi=\{z_j\}_{j=1}^m$ denotes the sampled minibatch.
For simplicity, we remove $\xi$
when the underlying minibatch is clear from context and write
$f_m(x_1,x_2)$.
\paragraph{Expected loss on the current population.}
The first term in~\eqref{eq:performative_decomp} can be estimated
without deploying the candidate $x_k+s$.
At iteration $k$, after collecting
$z_i\sim D(x_k)$, $i=1,\ldots,m_{\mathrm{ZO}}$, the first term in
\eqref{eq:performative_decomp} is estimated by $f_{m_{\mathrm{ZO}}}(x_k+s,x_k).$
Once these samples are collected, the empirical loss can be evaluated
at different candidate decisions $x_k+s$ without further deployments.

\paragraph{Population-response gradient.}
The second term in~\eqref{eq:performative_decomp} captures the effect
of the population response. Its derivative with respect to $s$ at
$s=0$ is
$\nabla_{x_2}R(x_k,x_k).$

For a pair $(x_1,x_2)$ and each coordinate
$i=1,\ldots,d$, let $e_i$ denote the $i$th standard basis vector and
let $\sigma>0$ be the finite-difference step. Using a base minibatch
from $D(x_2)$ and an independent minibatch from
$D(x_2+\sigma e_i)$, each of size $m_{\mathrm{ZO}}$, define
\begin{equation}
[g_2(x_1,x_2)]_i
=
\frac{
f_{m_{\mathrm{ZO}}}(x_1,x_2+\sigma e_i)
-
f_{m_{\mathrm{ZO}}}(x_1,x_2)
}{\sigma},
\qquad
i=1,\ldots,d.
\label{eq:est_1_2_formula}
\end{equation}
At iteration $k$, we set
$
g_{2,k}
:=
g_2(x_k,x_k)
$ to approximate $\nabla_{x_2} R(x_k,x_k).$

\paragraph{Local Approximation.}
Combining the empirical loss on the current population with the
response-gradient estimate gives the local approximation
\begin{equation}
m_k(s)
=
f_{m_{\mathrm{ZO}}}(x_k +s, x_k)
+
g_{2,k}^{\top}s
+
\frac{1}{2}s^\top H_k s
\label{eq:performative_TR_model}
\end{equation}

The matrix $H_k$ provides a quadratic correction.
Our analysis requires $H_k$ to be uniformly bounded, but does not
require it to accurately approximate the Hessian of $PR$. In practice,
one may simply choose $H_k=0$ or $H_k=\lambda I$ for a fixed
$\lambda\ge 0$.
\begin{assumption}
\label{assp:H_k_bound}
For every iteration $k = 0, 1, 2, \cdots, \| H_k\| \le \kappa_{H}$, where $\kappa_{H} > 0$.
\end{assumption}

Given a trust-region radius $\alpha_k>0$, PTRO approximately solves
$\min_{\|s\|\leq\alpha_k} m_k(s).
$

To specify the required solution accuracy, define the approximation gradient $v_k := \nabla m_k(0), $ So
\begin{equation}
v_k=\nabla_{x_1}f_{m_{\mathrm{ZO}}}(x_k,x_k)
    +g_{2,k} =
\frac{1}{m_{\mathrm{ZO}}}
\sum_{i=1}^{m_{\mathrm{ZO}}}
\nabla_x\ell(z_i^k;x_k)
+
g_{2,k}.
    \label{eq:v_k}
\end{equation}

We require a feasible step $s_k$ satisfying the
fraction-of-Cauchy-decrease condition
\begin{equation}
m_k(0)-m_k(s_k)
\geq
\frac{\kappa_{\mathrm{fcd}}}{2}\|v_k\|
\min\left\{
    \frac{\|v_k\|}{L_\ell+\kappa_H},
    \alpha_k
\right\},
\label{eq:fraction_cauchy_decrease}
\end{equation}
where $\kappa_{\mathrm{fcd}}\in(0,1]$ and $L_\ell$ is the
sample-wise loss smoothness constant from
Assumption~\ref{assp:Smoothness_in_loss_function}.
This condition ensures sufficient model decrease without requiring
an exact solution of the subproblem.

\subsection{Trust-Region Updates}

PTRO adjusts the
trust-region radius $\alpha_k$ by comparing predicted and observed risk reductions.

Let $x_k^+=x_k+s_k$ be the trial decision. We use empirical estimates
$f_{m_{\mathrm{ZO}}}(x_k,x_k)$ and $f_{m_{\mathrm{ZO}}}(x_k^+,x_k^+)$ of $PR(x_k)$ and $PR(x_k^+)$, respectively.
Estimating $f_{m_{\mathrm{ZO}}}(x_k^+,x_k^+)$ needs samples from the distribution $D(x_k^+)$
induced by the trial deployment. 

The acceptance ratio is
\vspace{-1.5em}
\begin{equation}
    \rho_k
    :=
    \frac{f_{m_{\mathrm{ZO}}}(x_k,x_k)-f_{m_{\mathrm{ZO}}}(x_k^+,x_k^+) + 2\epsilon_f}{m_k(0)-m_k(s_k)},
    \label{alg_eq:suffcient_decrease}
\end{equation}
where $\epsilon_f$ is the accuracy parameter of the performative risk estimator.
The term $2\epsilon_f$ relaxes the acceptance test to account
for errors in the two risk estimates.

If $\rho_k\geq\eta_1$, PTRO accepts the trial decision. Otherwise,
it keeps $x_k$ and shrinks the radius. Recall $v_k=\nabla m_k(0)$, following an accepted step,
the radius increases if $ \|v_k\|\geq\eta_2\alpha_k$ and decreases
otherwise. Here, $\eta_1,\eta_2\in(0,1)$,
$\gamma_{\mathrm{inc}}>1$, and
$\gamma_{\mathrm{dec}}\in(0,1)$ control acceptance and radius updates.

\subsection{PTRO}
\begin{assumption}[Feasibility of queried points]
All points queried by PTRO remain in $\mathcal X$; that is,
$x_k$, $x_k+s_k$, and $x_k+\sigma e_i$ belong to $\mathcal X$
for all $k$ and $i\in[d]$.
\label{assp:feasibility}
\end{assumption}

Assumption~\ref{assp:feasibility} is assumed throughout.
Algorithm~\ref{alg:trust_region} summarizes the procedure. The radius
adapts to the observed agreement between the local approximation and the
empirical risk reduction, so the update rule does not require
knowledge of the global smoothness constant of $PR$.

\paragraph{Sampling and parameter choices.}
Lemma~\ref{lemma:FO_construction} in
Appendix~\ref{subsec:Estimator_accuracy_guarantees_sample_complexity}
specifies choices of $\sigma$ and $m_{\mathrm{ZO}}$ that ensure the
desired response-gradient accuracy.

We reuse the same base minibatch of $m_{\mathrm{ZO}}$ samples from
$D(x_k)$ to construct
$f_{m_{\mathrm{ZO}}}(x_k+s,x_k)$,
$\nabla_{x_1}f_{m_{\mathrm{ZO}}}(x_k,x_k)$,
the base function estimate in $g_{2,k}$, and
$f_{m_{\mathrm{ZO}}}(x_k,x_k)$ in the acceptance test. So one iteration uses $N_{\mathrm{iter}}=(d+2)m_{\mathrm{ZO}}$ samples. This reuse does not introduce bias and our analysis does not
require independence.

\begin{algorithm}
\caption{PTRO}
\label{alg:trust_region}
\begin{algorithmic}[1]
\Require Initial decision $x_0$ and radius $\alpha_0>0$;
$\eta_1,\eta_2\in(0,1)$;
$\gamma_{\mathrm{inc}}>1$;
$\gamma_{\mathrm{dec}}\in(0,1)$;
risk-estimation accuracy $\epsilon_f>0$.
\For{$k = 0,1,2,\dots$}
    \State Collect $ m_{\mathrm{ZO}}$ samples from $D(x_k)$
    \State For $i = 1,\cdots, d,$ deploy $x_k +\sigma e_i$ and collect $m_{\mathrm{ZO}}$ samples
    \State Estimate $g_{2,k}$ using~\eqref{eq:est_1_2_formula}
    and choose $H_k$ satisfying
    Assumption~\ref{assp:H_k_bound}.
    
    \State Construct $m_k$ and $v_k$ using~\eqref{eq:performative_TR_model},~\eqref{eq:v_k} and compute $s_k$ satisfying~\eqref{eq:fraction_cauchy_decrease}.
    \State Set $x_k^+=x_k+s_k$ and deploy $x_k^+$, collect
    $m_{\mathrm{ZO}}$ samples from $D(x_k^+)$ to construct
$f_{m_{\mathrm{ZO}}}(x_k^+,x_k^+)$;.
     \State Compute $\rho_k$
    using~\eqref{alg_eq:suffcient_decrease}.
   \If{$\rho_k \ge \eta_1$}
    \State $x_{k+1} \gets x_k^{+}$ (accept step)
    \State $
    \alpha_{k+1} \gets
    \begin{cases}
        \gamma_{\text{inc}}\alpha_k, & \text{if }\|v_k\| \ge \eta_2 \alpha_k, \\
        \gamma_{\text{dec}}\alpha_k, & \text{if }\|v_k\| < \eta_2 \alpha_k .
    \end{cases}$
\Else
    \State $x_{k+1} \gets x_k$ (reject step)
    \State $\alpha_{k+1} \gets \gamma_{\text{dec}}\alpha_k$
\EndIf
\EndFor
\end{algorithmic}
\end{algorithm}

\section{Convergence results}
\label{sec:converge_complexity_result}
This section establishes the high-probability convergence and sample complexity of PTRO under the stated regularity in
Assumption~\ref{assp:PR_smooth}, the moment-bounded loss condition as~\eqref{eq:moment_bdd_loss}, and the bound on $H_k$ in
Assumption~\ref{assp:H_k_bound}. First, we state the required accuracy for estimators, then characterize the
stationarity tolerances compatible with the accuracy condition, and finally derive
the convergence and sample-complexity guarantees.

\subsection{Estimator Accuracy}

\paragraph{Assumptions.} These assumptions allow us to control the error of estimating risk and gradient.
\begin{assumption}[Sample-wise loss smoothness]
\label{assp:Smoothness_in_loss_function}
For every $z$, the loss $\ell(z;\cdot)$ is continuously differentiable
and $L_\ell$-smooth, i.e.,
$
\|\nabla_x\ell(z;x)-\nabla_x\ell(z;x')\|
\le
L_\ell\|x-x'\|,
\qquad
\forall x,x'\in\mathcal{X}.
$
\end{assumption}

\begin{assumption}[Bounded stochastic gradient variance]
    For all $x_1, x_2 \in \mathcal{X}$, the stochastic gradient with respect to $x_1$ satisfies
$\mathbb{E}_{z\sim D(x_2)}\!\left[\| \nabla \ell(z; x_1) - \nabla_{x_1} R(x_1, x_2)\|^2\right] \le \nu^2.$
\label{assp:gradient_var_bdd}
\end{assumption}
\begin{assumption}[Smoothness in the distribution argument]
\label{assp:nabla_x_2_R_lip}
For every $x_1\in\mathbb{R}^d$, the function $R(x_1,\cdot)$ is
continuously differentiable, and its gradient satisfies
$$\|\nabla_{x_2}R(x_1,x_2)-\nabla_{x_2}R(x_1,x_2')\|
    \leq L_{22}\|x_2-x_2'\|,
    \qquad \forall x_2,x_2'\in\mathcal{X},
$$
\end{assumption}

\paragraph{Estimator accuracy.}

For the risk estimator, define
$
e(x_1,x_2;\xi_f) := |f(x_1,x_2; \xi_f) - R(x_1,x_2)|
$ and we will drop the dependence on $\xi_f$ for brevity. 

For iteration $k$, let $X_k$ denote the random iterate generated by PTRO at iteration $k$,
with realization $x_k$, and let $G_{2,k}$ denote the random
response-gradient estimator, with realization $g_{2,k}$.

\added{Let
$E_k:=e(X_k,X_k),
E_k^+:=e(X_k^+,X_k^+),$
 let $\mathcal F_{k-1}$ denote the algorithmic history before
iteration $k$. Define
$
Y_k
:=
(E_k+E_k^+)
-
\mathbb E[E_k+E_k^+\mid\mathcal F_{k-1}].
$
We assume, almost surely,}
\begin{align}
\mathbb E[E_k+E_k^+\mid\mathcal F_{k-1}]
\le 2\epsilon_f, \quad
\mathbb E[|Y_k|^q\mid\mathcal F_{k-1}]\le 2^q\zeta_q,
\quad q\ge 2.
\label{eq:moment-bdd-oracle}
\end{align}


For the response-gradient estimator, we require
\begin{align}
\mathbb P\!\left(
\left.
\|G_{2,k}-\nabla_{x_2}R(X_k,X_k)\|
\le \epsilon_{g,2}
\,\right|\,\mathcal F_{k-1}
\right)
\ge 1-\delta_1
\label{eq:g_oracle}
\end{align}


Under the assumptions above, the estimator $f_{m_\mathrm{ZO}}(x,x),g_2(x,x)$ can achieve these accuracy levels with appropriate
batch sizes $m_{\mathrm{ZO}}$ and finite-difference radius $\sigma$, as shown in 
Lemma~\ref{lemma:sample_cplx_zero_order_erre_bdd_moment} and~\ref{lemma:FO_construction}. 
\subsection{High-Probability Convergence}
We now establish the high-probability convergence of PTRO.
The argument has three ingredients: sufficiently accurate
(\emph{true}) iterations occur with uniformly positive
probability; on such iterations, the trust-region mechanism guarantees
progress ; and
the resulting progress accumulates sufficiently often to reach an
$\epsilon$-stationary point.
\paragraph{True iteration}
We first formalize the event that iterations  are accurate.
\begin{definition}[True iteration]
\label{def:true_iteration}
An iteration $k$ is called a true iteration if
\begin{align*}
\left\|
\nabla_{x_1}\widehat R_k(X_k,X_k)
-\nabla_{x_1}R(X_k,X_k)
\right\|
&\le \epsilon_{g,1},\;
\left\|
G_{2,k}-\nabla_{x_2}R(X_k,X_k)
\right\|
\le \epsilon_{g,2},\;
E_k+E_k^+
\le 2\epsilon_f,
\end{align*}
where
$E_k:=|f_k-PR(X_k)|,
\quad
E_k^+:=|f_k^+-PR(X_k^+)|.$

\end{definition}

Let $\epsilon_g:=\epsilon_{g,1}+\epsilon_{g,2}.$
On a true iteration, the corresponding realizations satisfy $\|v_k-\nabla PR(x_k)\|
    \le \epsilon_g,$ and $|(f_k-f_k^+)
        -
        \bigl(PR(x_k)-PR(x_k^+)\bigr)|
    \le 2\epsilon_f.$ Thus, a true iteration means both the gradient of local approximation and the
acceptance test are reliable.

Recall $\mathcal F_{k-1}$ is the history before iteration $k$, and let $I_k := \mathbf{1}\{\text{iteration } k \text{ is true}\}$.

\begin{lemma}[Probability of a true iteration]
\label{lemma:true_iteration_probability}
    Suppose Assumption~\ref{assp:gradient_var_bdd}, the moment condition
\eqref{eq:moment_bdd_loss}, and the estimator condition
\eqref{eq:g_oracle} hold.  Then, for all $k$,
\[
\mathbb{P}(I_k=1\mid\mathcal{F}_{k-1})
\ge
1-\frac{\nu^2}{m_{\mathrm{ZO}}\epsilon_{g,1}^2}
-\delta_1
-\frac{2L_q^{2/q}}{m_{\mathrm{ZO}}\epsilon_f^2}
=:p.
\]
In particular, if the sample sizes are chosen such that
$p>0$, the probability of a true iteration is uniformly
bounded below by $p$.
    \label{lemma:p_lower_bd}
\end{lemma}
As a result, batch size and estimator accuracies determine the probability lower bound $p$ of true iterations.

\paragraph{Admissible stationarity tolerance}
To guarantee progress toward an $\epsilon$-stationary point, we require
two conditions to regulate $\epsilon_f$ and $\epsilon_g$. First, the gradient-estimation error must be small
relative to $\epsilon$. Second, true iterations must occur 
frequently.

To quantify the latter requirement, define
    $u:=
    -\frac{\ln\gamma_{\mathrm{inc}}}
           {\ln\gamma_{\mathrm{dec}}},$
and let
$h(\epsilon)=C_h\epsilon^2,$ and 
$
    p_u
    :=
    \frac{1}{u+1}
    +
    \frac{4u\epsilon_f}{h(\epsilon)}.
$
Here, $p_u$ is the minimum frequency of true iterations required by the
trust-region progress analysis, accounting for both radius updates and
risk-estimation error. For simplicity, we assume throughout that $u$ is a positive integer.

The following proposition gives explicit condition for estimator accuracy.
\begin{proposition}
\label{prop:lower_bd_for_eps}
Suppose $p>\frac{1}{u+1},$ choose
$0<\eta_g<
    \frac{\kappa_{\mathrm{fcd}}(1-\eta_1)}
         {2+\kappa_{\mathrm{fcd}}(1-\eta_1)}.
$
If
$$
    \epsilon_g<\eta_g\epsilon,
    \qquad
    \epsilon_f<
    \frac{C_h}{4u}
    \left(
        p-\frac{1}{u+1}
    \right)\epsilon^2,
$$
then $p>p_u$, and $\epsilon$ is admissible for the convergence
guarantee.
\end{proposition}

Here, $\eta_g$ specifies how small the gradient-estimation error must be
relative to the target tolerance. 
Thus, it requires the gradient-estimation error scale as
$\epsilon_g=O(\epsilon)$ and the risk-estimation error scale as
$\epsilon_f=O(\epsilon^2)$, which in turn requires corresponding increases in the batch size.
We discuss the sufficient per iteration sample complexity in Section~\ref{subsec:sample_cplx}.

\paragraph{Convergence guarantee.}
We now state the high-probability convergence result. Let
$\bar\alpha:=C_\alpha\epsilon,
$
where $C_\alpha>0$ depends only on the smoothness and algorithm
parameters. Explicit expressions of $C_\alpha$ and $C_h$
are given in the appendix.

\begin{theorem}
[High-probability convergence] Suppose the stated regularity as Assumption~\ref{assp:PR_smooth} and estimator-accuracy conditions~\eqref{eq:moment-bdd-oracle} and~\eqref{eq:g_oracle} hold.

Choose the estimator accuracies so that the target $\epsilon$ is
admissible.

Let $X_k$ denote the random iterate generated by PTRO with initial
trust-region radius $\alpha_0>$\added{$\bar \alpha$}. 

Define $Z_0:= PR(X_0) - PR^*$, and $B_\epsilon= \frac{uZ_0}{h(\epsilon)} + \frac{u}{u+1} \max\left\{
0,\,
\left\lceil
\frac{\ln(\alpha_0/\bar\alpha)}
{\ln(1/\gamma_{\mathrm{dec}})}
\right\rceil
\right\}.$ 

Then for any $s \ge 0$, and
$\widehat p\in(p_u+\frac{us}{h(\epsilon)},p)$, if
$
    t>
    \frac{B_\epsilon}
         {\widehat p-p_u-us/h(\epsilon)},$
then:
\begin{equation}
\small{
   \mathbb{P}\left\{
\min_{0\le k\le t-1}\|\nabla PR(X_k)\|\le \epsilon \right \} \ge 1 - \exp \left(-\frac{(p - \hat{p})^2}{2\max\{p, 1-p\}^2}t \right) - \delta_t(s)}
\label{eq:prop_converge} 
\end{equation}
where 
 $ \delta_t(s)
=
\exp\!\left(
-\frac{s^2}{2(q+2)^2 e^q \zeta_2}\, t
\right)
+
\frac{
\left(1 + \frac{2}{q}\right)^q \, 2^q \zeta_q
}{
s^q \, t^{\,q-1}
}, \text{ with }\zeta_2 = \frac{L_q^{2/q}}{m_{\mathrm{ZO}}}.$
\label{thm:convergence}
\end{theorem}
The theorem separates two sources of failure: too few true iterations
and excessive accumulated risk-estimation error. The first produces
the exponential term, and the second produces $\delta_t(s)$.


\begin{remark}[Iteration complexity]
Theorem~\ref{thm:convergence} gives a high-probability iteration complexity of order $O(\epsilon^{-2})$. Because $h(\epsilon) = \Theta(\epsilon^2)$, then define $t_s:=\frac{R}{\hat p-p_u-\frac{s}{h(\epsilon)}}$, we have $t_s = O(\epsilon^{-2})$. 
\end{remark}


\subsection{Sample complexity}
\label{subsec:sample_cplx}
We now specify the sampling requirements of each iteration for stationarity
tolerance $\epsilon$. 
\begin{proposition}[Per-iteration sample complexity]
\label{prop:sample_complexity}
Suppose the moment-bounded loss condition holds for a fixed
$q\geq2$, together with the stated smoothness and
stochastic-gradient variance assumptions.
For a target $\epsilon\in(0,1)$, the estimator parameters can be
chosen so that $p>p_u(\epsilon)$, using
$$m_{\mathrm{ZO}}
=\Theta\left(d^2\epsilon^{-4}\right).
$$
Recall
$N_{\mathrm{iter}}
    =(d+2)m_{\mathrm{ZO}}.,$ the number of samples per iteration satisfies
$
    N_{\mathrm{iter}}
=(d+2)m_{\mathrm{ZO}}
=O\left(d^3\epsilon^{-4}\right).
$
The hidden constants depend on the moment, smoothness, variance,
and algorithm parameters.
\end{proposition}

\paragraph{Dimension dependence}
The dimension dependency in sample complexity has two sources:(1) $O(d)$ perturbed deployments for coordinate-wise
finite-difference estimation of the population-response
gradient;
(2) $O(d^2)$ for minibatch
size per deployment, sufficient to control both
finite-difference bias and sampling variance and to
satisfy the conditional high-probability full-gradient
accuracy requirement.

Randomized finite-difference estimators could reduce the number of perturbed deployments per iteration from $O(d)$ to $O(1)$. But this does not necessarily eliminate the dimension dependency in the overall sample complexity, as randomized gradient estimators generally introduce dimension-dependent variance. 
Notably, the batch sizes are sufficient conditions, rather than necessary sampling requirements.
Our experiments in~\ref{appdx_subsec:effects_of_dim_budget} indicate that this dependence may be conservative in
practice: quadratic scaling already removes most of the observed gap across dimensions on the tested problem.

\paragraph{Possible improvements.}
There are several
possible directions for reducing this cost.

First, in many applications the population may not respond separately
to every model parameter. Instead, the induced distribution may depend
on the high-dimensional decision $x\in\mathbb{R}^d$ only through a
lower-dimensional deployed quantity $
a=h(x)\in\mathbb{R}^r,
 r\ll d.
$ For example, in strategic classification, individuals may respond only
through a small subset of manipulable or decision-relevant quantities,
rather than through all parameters. In such
settings, it is potentially replacing the dependence
on $d$ by dependence on $r$.

Second, the decoupled-risk formulation may allow reuse of historical
deployment data. For example, nearby historical deployments could be used
to construct a local interpolation or regression model of
$R(x_k,\cdot)$, potentially reducing the need to collect fresh samples
from all $d$ coordinate perturbations at every iteration. A rigorous
analysis of such reuse would require additional conditions on the
coverage and geometry of the historical deployment points, and we
leave this direction for future work.

\section{Experiments}
\subsection{Strategic classification with Heterogeneous Responses}
\label{subsec:kaggle_exp}
We use the Kaggle \emph{Give Me Some Credit} credit scoring dataset~\citep{givemesomecredit2012}, commonly used in previous work ~\citep{perdomo2020performative,mendler2020stochastic,brown2022performative,wang2023network}, and the the \emph{ACSIncome} task from Folktables~\citep{ding2021retiring}, which predicts
whether an individual's income exceeds \$50k from American Community Survey (ACS)
records~\citep{acs_pums}

We use logistic loss and consider  three
response mechanisms: (1) \textbf{Linear.} All individuals follow the same linear response rule.(2) \textbf{Mixture-linear (MixL).}
The population is divided into two groups. The groups share the linear response but respond in opposite directions. (3) \textbf{Mixture-nonlinear (MixN).}
We keep the two  groups and replace the
linear response with a nonlinear rule. These three response mechanisms are designed to progressively violate the structural assumptions used by response-model-based baselines.

The learner observes only the loss and receives no group information. 
We compare PTRO with Two-stage, PerfGD, DSA-PGM, DFO$(\lambda)$, and GZO. All methods are given the same total sample budget. 

\begin{figure}[t]
    \centering
    \begin{subfigure}{\linewidth}
        \centering
        \includegraphics[width=\linewidth]{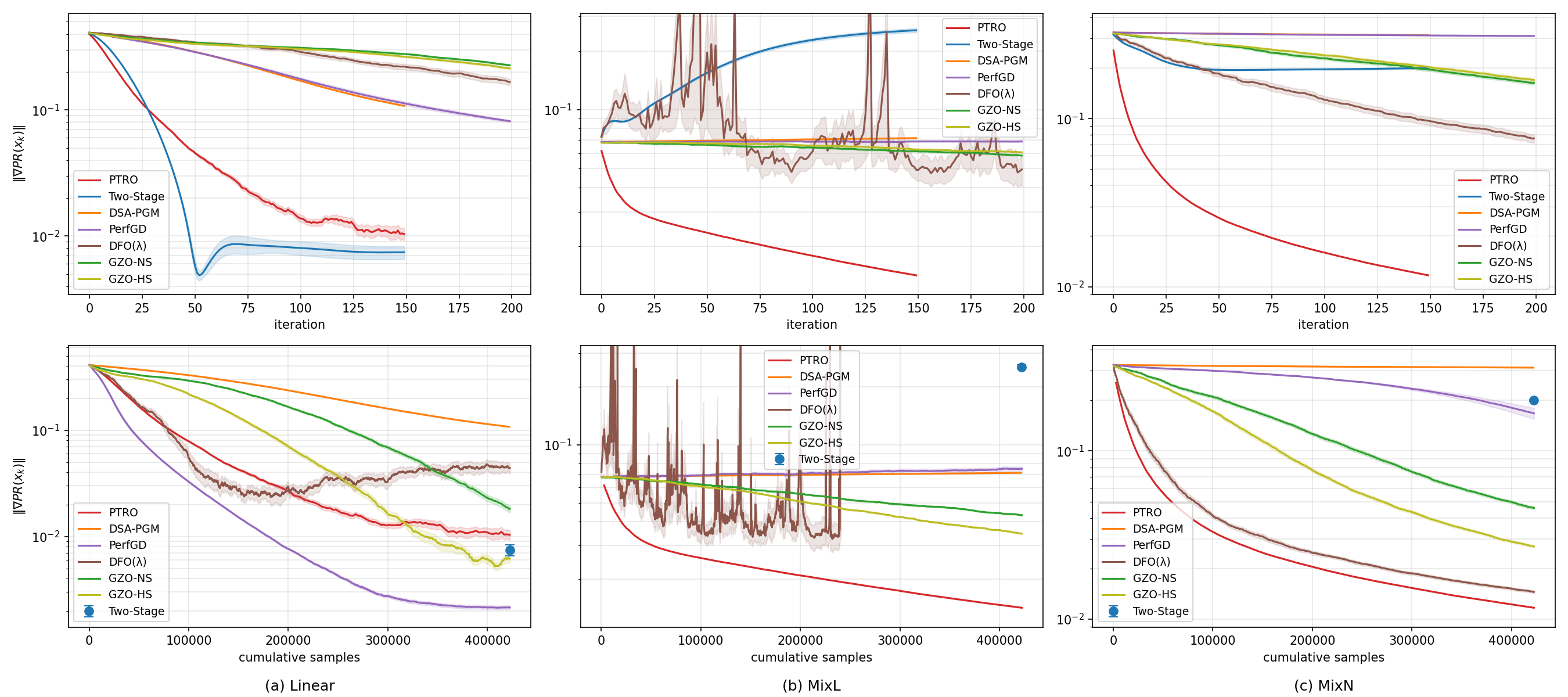}
        \caption{Give Me Some Credit}
        \label{fig:comparison_gmsc}
    \end{subfigure}

    \vspace{0.5em}

    \begin{subfigure}{\linewidth}
        \centering
        \includegraphics[width=\linewidth]{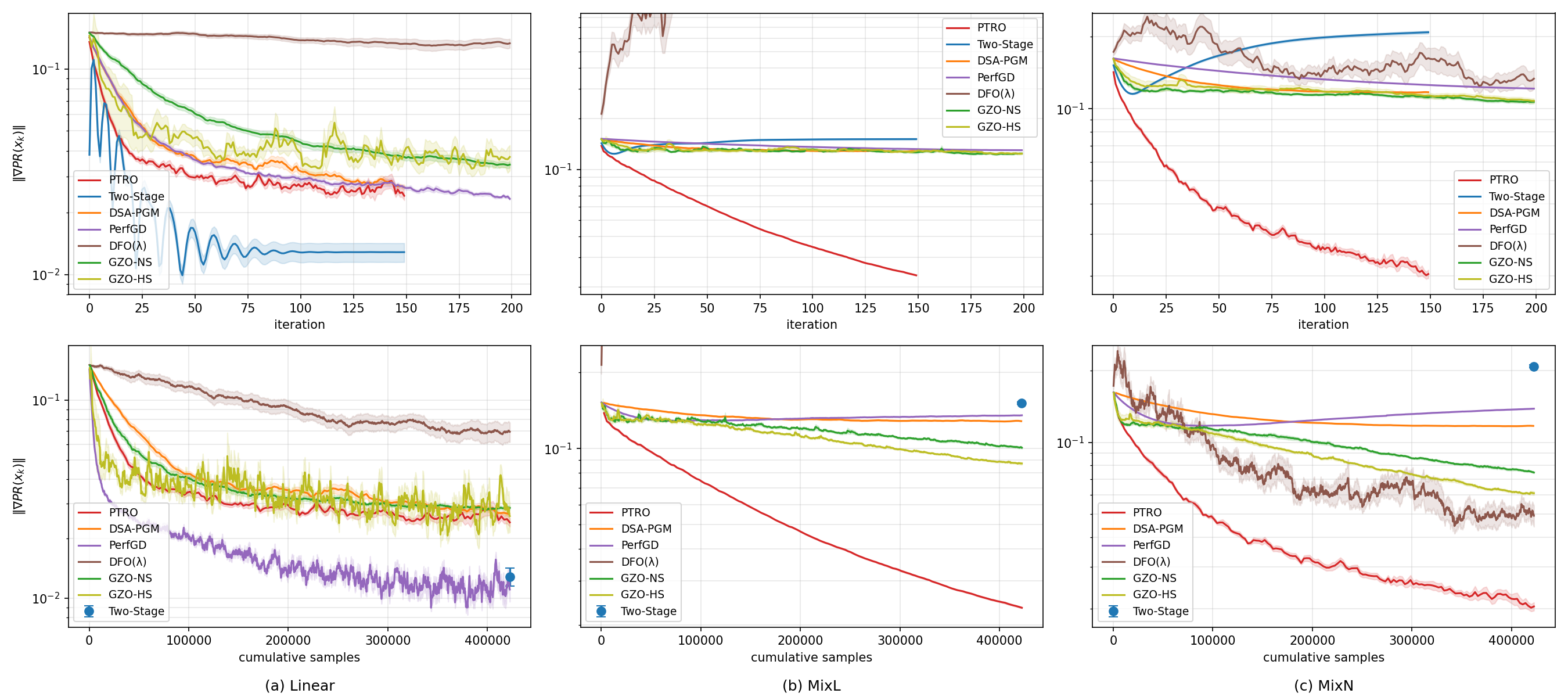}
        \caption{ACSIncome}
        \label{fig:comparison_acs}
    \end{subfigure}
    \caption{Comparison under the Linear, MixL,
    and MixN responses. All methods
    use the same total sample budget and are evaluated on the same fixed held-out
    population.}
    \label{fig:comparison_methods}
\end{figure}

Figure~\ref{fig:comparison_methods} compares the methods under three population-response mechanisms.

Under the homogeneous linear response, where the response model is correctly
specified, model-based methods can be more sample-efficient than PTRO. On Give Me
Some Credit, PerfGD reaches the lowest stationarity level, followed by Two-Stage,
both below PTRO; on ACSIncome, PerfGD and Two-Stage likewise attain lower levels
than PTRO under the same cumulative sample budget. These results illustrate the benefit of exploiting a
correctly specified response model. DFO$(\lambda)$ makes progress on Give Me Some
Credit but fluctuates and does not improve on ACSIncome, consistent with the high
variance of its randomized one-point gradient estimator.

Under the heterogeneous linear (MixL) and nonlinear (MixN) responses, PTRO
consistently reduces the stationarity measure and achieves the lowest level under
the same sample budget on both datasets. In contrast, the response-model-based
methods stagnate or even deteriorate; for example, Two-Stage ends at a higher
stationarity level than its starting point under MixL. DFO$(\lambda)$, GZO-NS and GZO-HS, which
also do not require a parametric response model, make some progress but
remain less effective than PTRO. The reason is their search direction is guided by the direct-effect gradient
$\nabla_{\theta_1} R$, but in our settings $\nabla\mathrm{PR}$ is dominated by the
response term ($\|\nabla\mathrm{PR}-\nabla_{x_1}R\|/\|\nabla\mathrm{PR}\|\approx 1$),
and the squared cosine between the guide and $\nabla\mathrm{PR}$ stays below $0.05$. The guided component of the
update therefore contributes almost nothing. DFO($\lambda$)'s instability may be related to the variance
of its randomized one-point gradient estimator.

Overall, the results illustrate the tradeoff between structural assumptions
and robustness to response misspecification: exploiting a well-specified
response model can improve sample efficiency, whereas PTRO remains effective
as the population response becomes poorly captured by the assumed model.




Additional experiments systematically vary response heterogeneity. Appendix~\ref{appdx_subsec:Robustness to Population Heterogeneity} continuously increases the fraction of the population following a different response mechanism, while Appendix~\ref{appdx_subsec:effect_of_response_distribution} separately varies the magnitude and number of heterogeneous response types. These experiments show that PTRO remains stable as population heterogeneity increases and that the magnitude has a larger effect on optimization than the number of distinct response types. Ablations demonstrate the benefits of exploiting the known loss and adaptively adjusting the trust-region radius. Experiments across dimensions and sampling budgets suggest that the sample-complexity bound in Proposition~2 is conservative in practice, while sensitivity experiments show that PTRO is generally robust to its hyperparameter choices; see Appendices~\ref{appdix:ablation_setup}, \ref{appdx_subsec:effects_of_dim_budget}, and \ref{appdx_subsec:sensitivity_exp}.

\section{Conclusion}

This work studies performative optimization when the population response cannot
be reliably represented by a prespecified model. Our results suggest that
explicit response modeling is not necessary for obtaining principled
optimization guarantees: by exploiting the known loss and estimating only the
effect of population response, PTRO converges to stationary points without
assuming a known parametric form for the response.

More broadly, relaxing assumptions on the population-response distribution
introduces a tradeoff between structural knowledge and sampling efficiency.
Without a prescribed response model, PTRO must learn the effect of population
response through additional deployments, leading to a dimension-dependent
sampling cost. Moreover, exploratory deployments may affect exposed individuals
even when trial steps are rejected. Future work could explore more deployment-efficient
estimators, historical data reuse, and safety-constrained exploration to reduce
sampling costs and mitigate deployment risks without requiring a fully specified
response model.

\clearpage
\bibliographystyle{plainnat}
\bibliography{reference}

\clearpage

\begin{center}
    \Large \textbf{Appendix}
\end{center}
\listofappendices
\clearpage
\appendix

\section{Background}
\addcontentsline{apc}{section}{\protect\numberline{\thesection}Background}
\label{appdx:background}

\subsection{Related Work}
\addcontentsline{apc}{subsection}{\protect\numberline{\thesubsection}Related Work}
\label{appdx:related_work}
\paragraph{Performative Optimization }
Performative prediction studies decision-making problems in which the
data distribution changes in response to the deployed decision
\citep{perdomo2020performative}. Two central solution concepts are
\emph{performative stability} and \emph{performative optimality}.
A performatively stable point is a decision that is optimal for the
distribution induced by deploying that same decision. Formally,
$x_{\mathrm{PS}}$ satisfies
\begin{equation}
    x_{\mathrm{PS}}
    \in
    \arg\min_x R(x,x_{\mathrm{PS}}).
\label{eq:performative_stable}
\end{equation}

where the distribution is held fixed at the one induced by the current
decision. Many works study convergence to
performatively stable points
\citep{perdomo2020performative,brown2022performative,
drusvyatskiy2023stochastic, zhong2025nonlinear}, or their first-order counterparts
\citep{li2024stochastic}. Several works further consider approximate notions of performative
stability. For example,~\citet{farina2026stability} measure stability
through the best-response risk gap
$
R(x^\star,x^\star)-\min_x R(x,x^\star)\le\epsilon,
$
while~\citet{anagnostides2026computational} consider the first-order
relaxation
$
\left\langle
x-x^\star,\nabla_{x_1}R(x^\star,x^\star)
\right\rangle
\ge -\epsilon,
 \forall x\in\mathcal X.
$
All of these notions evaluate optimality under the induced
distribution fixed at $D(x^\star)$.

However, a
performatively stable point can be arbitrarily far from the
performatively optimal solution~\citep{miller2021outside}, because it does not account for how a
change in the decision changes the population. This
motivates directly optimizing the performative risk, whose gradient contains both the direct effect of changing the decision
and the indirect effect through distribution shift:
$
\nabla PR(x)
=
\nabla_{x_1}R(x,x)
+
\nabla_{x_2}R(x,x).
$

Some works study performatively optimal solutions
\citep{miller2021outside,ray2022decision,lin2023plug},
sublinear regret~\citep{chen2023performative}, and
stationarity guarantees for performative risk minimization
\citep{izzo2021learn,liu2026stochastic}.
In general, $PR(\cdot)$ can be nonconvex even when the loss function
and the distribution map $D(\cdot)$ satisfy favorable
regularity conditions, as illustrated by the following example.
\begin{example}
    
\label{exp:nonconvex}
Consider the setting
$ z_x~\overset{d}{=}~z_0~+~\rho \tanh(x)$ with $z_0 \sim \mathcal{N}(0,1).$ 
Let
$ \ell(z;x)=-zx+\frac{\gamma}{2}x^2, \text{ with }\gamma>0. $ The performative risk is
$ \mathrm{PR}(x)
=-\rho x\tanh(x)+\frac{\gamma}{2}x^2, $ which has $PR''(0)=
\gamma-2\rho<0$. So
when $ \rho>\frac{\gamma}{2}, $ it is non-convex.
Notably, the $\ell(z;x)$ is $\gamma$-strongly convex in $x$, and jointly $\beta$-smooth with
$ \beta=\frac{\gamma+\sqrt{\gamma^2+4}}{2}, $ and the distribution map has sensitivity at most $\rho$ in Wasserstein-1 distance.
\end{example}
Therefore, this example motivates studying performative risk minimization without
assuming global convexity. Accordingly, our target is to seek an $\epsilon$-stationary point of the full
performative risk: $\|\nabla PR(x)\|\le\epsilon.$

Another line of work considers distributionally robust performative
optimization~\citep{xue2024distributionally,jia2026distributionally},
where the response distribution itself is allowed to vary within an
uncertainty set. A representative objective takes the form
$
\operatorname{DRPR}(x)
=
\sup_{\widetilde D\in\mathcal U(D)}
\mathbb E_{Z\sim\widetilde D(x)}
[\ell(Z;x)].
$
This formulation addresses robustness to ambiguity in the response
distribution, whereas our focus is on directly optimizing the
performative risk under an unknown response map using stochastic
zeroth- and first-order information.
\paragraph{Derivative-free optimization and gradient estimation}
Derivative-free optimization~\citep{flaxman2004online,ghadimi2013stochastic,ji2019improved,nesterov2017random,gasnikov2017stochastic} is particularly relevant to performative prediction since $D(\cdot)$ is unavailable. But treating $PR$ as a pure black-box is unnecessary because $\ell$ is often known and $\nabla_{x_1}R(x,x) = \mathbb E_{z\sim D(x)}[\nabla_x\ell(z;x)] $ can be estimated directly from samples. 

To estimate $\nabla_{x_2}R$, there are several derivative-free estimators, including finite differences, interpolation-based estimators, and randomized smoothing. 
Coordinate finite differences are a standard choice in derivative-free optimization and have also been used in closely related settings~\citep{fang2026high,shi2022adaptive}.
\citet{berahas2022theoretical} provide a theoretical and empirical comparison of these alternatives. For large dimension $n$, all the compared methods have the same order-of-magnitude dependence on $n$, but that the constants in the sample bounds for smoothing methods can be larger than for deterministic methods such as finite differences. We therefore use coordinate-wise finite differences as a simple and representative choice rather than claiming it is
optimal.


\paragraph{Stochastic Trust-Region Methods}
Trust-region (TR) methods are particularly well suited to the deployment
structure of performative optimization. A change in the decision changes
the population distribution, so taking an unrestricted update
based on an inaccurate local approximation may expose the population to a
decision whose improvement has not been validated. TR methods
adds an explicit validation
stage: the proposed update is first restricted locally and is retained
only if sampled feedback supports the predicted improvement.

Stochastic TR methods have been developed for optimization with probabilistically accurate models and noisy function evaluations~\citep{gratton2018complexity,blanchet2019convergence,cao2024first,jin2025sample,scheinberg2025stochastic}. These works establish convergence by showing that sufficiently accurate iterations occur with positive probability and that successful iterations yield sufficient decrease. PTRO builds on this general framework, but the performative setting introduces a different local approximation structure:  we incorporate a decoupled gradient estimation, we consider the first-order estimator, whose error is bounded as $\epsilon_g + r(\alpha)$, where $r(\alpha)$ captures the stochastic gradient error, and $\epsilon_g$ represents the irreducible bias. Building on this model, we further derive explicit sample complexity bounds following~\cite{jin2025sample}. 

\subsection{Comparison with Existing Methods}
\addcontentsline{apc}{subsection}{\protect\numberline{\thesubsection}Comparison with Existing Methods}

First, Table~\ref{tab:assumption_comparison} compares conditions used in existing methods. Methods such as Two-stage, Plug-in, PerfGD, and DSA-PGM assume that the population response can be represented by a known family, with only its parameters or parameter map left unknown. This can make the response easier to estimate, but ties the method to that modeling choice. PTRO takes a different route. Rather than specifying how $D(x)$ is parameterized, it treats the induced distribution as an unknown object and imposes regularity only on quantities needed for optimization, such as moments of the loss and smoothness of the decoupled risk. Hence the same analysis can cover an unknown parametric family, an unknown mixture response, or a response distribution with no fixed finite-dimensional parameterization, provided these regularity conditions hold.

DFO$(\lambda)$ is also nonparametric in this sense. Its analysis
obtains tractability through regularity assumptions on the
decision-dependent stochastic process, including smooth dependence of the induced distribution on the decision and boundedness of the loss. In the Markovian
setting, it additionally assumes geometric mixing of the response
process.

\begin{table}[t]
\centering
\caption{
Comparison of structural and regularity assumptions in performative
optimization. 
}
\label{tab:assumption_comparison}

\tiny
\setlength{\tabcolsep}{3pt}
\renewcommand{\arraystretch}{1.15}

\begin{tabular}{
    @{}
    p{0.11\linewidth}
    p{0.20\linewidth}
    p{0.23\linewidth}
    p{0.20\linewidth}
    p{0.21\linewidth}
    @{}
}
\toprule
Method
&
\shortstack{Response\\parameterization}
&
Response regularity
&
Statistical control
&
Loss / risk regularity
\\
\midrule

PTRO
&
None
&
Lipschitz $\nabla_{x_2}R$
&
Finite $q$th loss moment;
bounded stochastic-gradient variance
&
Smooth loss and
 risk
\\

DFO$(\lambda)$
\citep{liu2023two}
&
None
&
\shortstack[l]{Controlled Markov response;\\
TV-Lipschitz dependence and\\
geometric mixing}
&
Bounded loss
&
Smooth risk
\\

Guided-ZO
\citep{hikima2025guided}
&
None
&
$W_1$-Lipschitz dependence
of $D(x)$
&
Bounded objective-value variance
&
Lipschitz loss;
smooth risk
\\

Two-stage
\citep{miller2021outside}
&
Location family
$z_x \overset{d}{=} z_0+\mu x$, unknown $\mu$
&
Response determined by
$\mu$
&
Sub-Gaussian $z_0$
&
Smooth, strongly convex loss
\\

Plug-in
\citep{lin2023plug}
&
assumes an atlas (a known family of candidate response maps)
&
Atlas regularity in
TV / $W_1$
&
Sub-Gaussian /
sub-exponential control
&
Bounded or Lipschitz loss
\\

PerfGD
\citep{izzo2021learn}
&
Known density family
$p(z;f(x))$, unknown $f$
&
$f$ has bounded first and second derivatives
&
Bounded loss, bounded distribution-parameter estimation error
&
Smooth loss and risk
\\

DSA-PGM
\citep{hamaguchi2026adaptive}
&
Unknown response map $\beta(x)\in\mathcal B$, with known $\mathcal B$
&
Differentiable parameter map $\beta(x)$ with Lipschitz Jacobian
&
Finite-moment of data,
bounded parameter-estimation MSE
&
Smooth loss with
polynomial growth; smooth risk (derived jointly with response regularity)
\\

\bottomrule
\end{tabular}
\end{table}

Second, we list existing methods for finding performatively optimal point or the first order stationary point and compare them with PTRO in details.

\paragraph{Two-stage algorithm}
\citet{miller2021outside} consider a location-family response model,
$$
z_x \sim D(x),
\qquad
z_x \overset{d}{=} z_0+\mu x,
$$
where $\mu$ is an unknown parameter and $z_0\sim D_0$.
Their method proceeds in two stages. In the first stage, it deploys
random decisions $x_i\sim\mathcal N(0,I_d)$, observes
$z_i\sim D(x_i)$, and estimates $\mu$ by ordinary least squares.
It then collects an additional set of samples from the base
distribution $D_0$ by repeatedly deploying the zero decision,
since $D(0)=D_0$ under the location-family model.

In the second stage, treating $\hat{\mu}$ as the true response
parameter, the algorithm uses the base samples
$z_j\sim D_0$, $j=n+1,\ldots,2n$, to construct the proxy objective
$$
\hat{x}_{\mathrm{PO}}
=
\arg\min_{x\in\mathcal X}
\frac{1}{n}
\sum_{j=n+1}^{2n}
\ell\!\left(z_j+\hat{\mu}x;\,x\right).
$$
Thus, $z_j+\hat{\mu}x$ simulates a sample from the estimated
distribution map at the candidate decision $x$.

\paragraph{Plug-in performative optimization.}
\citet{lin2023plug} consider a more general parametric family
\[
\mathcal D_B:=\{D_\beta:\beta\in B\},
\qquad
B\subseteq\mathbb R^{d_\beta},
\]
to model the unknown response map $D(\cdot)$, and define
$
PR_\beta(x)
=
\mathbb E_{z\sim D_\beta(x)}[\ell(z;x)].
$

Their method similarly first collects deployment data by sampling
decisions $x_i$ and observing
$z_i\sim D(x_i)$. It then fits the response model,
$
\hat\beta
=\hat{Map}
\bigl((x_1,z_1),\ldots,(x_n,z_n)\bigr),
$
and finally computes the plug-in performative optimal point
$$
\hat x_{\mathrm{PO}}
=
\arg\min_{x\in\mathcal X}
\mathbb E_{z\sim D_{\hat\beta}(x)}[\ell(z;x)].
$$

\paragraph{PerfGD}
\citet{izzo2021learn} assume that $D(x)$ has a continuously differentiable density $p(z; f (x))$, where the functional form of $p(z; w)$ is known and the quantity $f (x)$ can be easily estimated from a sample drawn from $D(x)$. The algorithm first performs $H$ iterations of Repeated empirical gradient descent(REGD), 
$$ x_{t+1} = x_t - \eta\,\hat{g}(x_t),
   \qquad
   \hat{g}(x_t) = \frac{1}{m}\sum_{i=1}^{m}\nabla_x\ell(z_i;x_t).
$$
to collect a history of deployed decisions and their induced responses.

After initial $H$ iteration, the algorithm switches to gradient descent using an estimate of the full performative gradient. At iteration $t$, it first estimates the response parameter $f(x_t)$ from samples drawn from $D(x_t)$ and uses the most
recent $H$ pairs
$
\{(x_j,\hat f(x_j))\}_{j=t-H+1}^{t}
$
to estimate $\nabla f(x_t)$.

The estimation with the known density
model $p(z;f(x))$ gives an estimate of the distribution-shift component
$\nabla_{x_2}R(x_t,x_t)$. PerfGD then adds this term and the direct
loss-gradient estimate $\hat g_1(x_t)$ to form a full performative gradient estimation, which is used to update $x_t$.

The result is deterministic under bounded
response-parameter estimation error.

\paragraph{Distribution-Shift-Aware Projected Gradient Method (DSA-PGM)}
\citet{hamaguchi2026adaptive}
similarly assume that the response distribution belongs to a known
family with unknown parameter mapping $\beta(x).$

At iteration $t$, the algorithm deploys the current decision
$x_t$ together with coordinate-wise perturbations
$x_t\pm\delta e_i$, $i=1,\ldots,d$, and collects samples from the
corresponding response distributions.
Using these samples, it first estimates the distribution parameters
$\hat\beta_t$ and $\hat\beta_{t,i}^{\pm}$ (corresponding with perturbed $x_t$), and then approximates the
Jacobian of the response parameter map by centered finite differences,
$$
\widehat{\frac{\partial \beta}{\partial x}}(x_t)
=
\left[
\frac{\hat\beta_{t,1}^{+}-\hat\beta_{t,1}^{-}}{2\delta},
\ldots,
\frac{\hat\beta_{t,d}^{+}-\hat\beta_{t,d}^{-}}{2\delta}
\right].
$$
The estimated Jacobian, together with the known density model
$p(z;\beta)$, is then used to construct the distribution-shift
component of the performative gradient. 
The full
performative gradient is formed by the component and a direct loss-gradient component, which is used to update $x_t$.

And the result is in expectation: $\frac{1}{T}\sum_{t=0}^{T-1} \mathbb{E}[\|G_\alpha(x)\|^2] \le \epsilon^2$, where
$\mathcal G_\alpha(x)=\alpha^{-1}
\!\left(x-\Pi_{\mathcal X}(x-\alpha\nabla PR(x))\right)$
is the projected-gradient mapping. 

\paragraph{DFO$(\lambda)$}\cite{liu2023two}  focuses on a generalization of our setting that induced distribution evolves according to a controlled Markov chain, such as the same group of customers gradually adjust their behavior under a deployed decision. This method is a fully derivative-free method that accesses only loss evaluations. 

At outer iteration $k$, DFO$(\lambda)$ samples a random direction
$u_k$ and repeatedly deploys perturbed decisions
\[
    \tilde x_k^{(m)}
    =
    x_k^{(m)}+\delta_k u_k,
    \qquad m=1,\ldots,\tau_k.
\]
After each deployment, it observes a new Markov state
$
    Z_k^{(m)}
    \sim
    T_{\tilde x_k^{(m)}}
    \bigl(Z_k^{(m-1)},\cdot\bigr),
$
and forms the one-point zeroth-order gradient estimate
\[
    g_k^{(m)}
    =
    \frac{d}{\delta_k}
    \ell\!\left(\tilde x_k^{(m)};Z_k^{(m)}\right)u_k.
\]
The decision is then updated within the inner loop as
\[
    x_k^{(m+1)}
    =
    x_k^{(m)}
    -
    \eta_k
    \lambda^{\tau_k-m}
    g_k^{(m)},
\]

\paragraph{Guided-ZO}
\cite{hikima2025guided} considers stochastic nonconvex
optimization with an unknown decision-dependent
distribution, without assuming a parametric response model.
Unlike conventional zeroth-order methods, Guided-ZO
exploits the known loss gradient to guide the random
perturbation directions.

At iteration $k$, it first estimates the direct loss-gradient
component using samples from $D(x_k)$:
\[
    s_k
    =
    \frac{1}{n}
    \sum_{i=1}^{n}
    \nabla_x \ell(x_k;z_k^i),
    \qquad z_k^i\sim D(x_k).
\]
It then constructs a covariance matrix
\[
    \Sigma_k
    =
    \frac{\alpha}{d}I_d
    +(1-\alpha)h_kh_k^\top,
    \qquad
    h_k=\frac{s_k}{\|s_k\|},
\]
where $h_k=0$ if $s_k=0$.
A random direction $u_k\sim\mathcal N(0,\Sigma_k)$
is sampled, and two perturbed decisions
$x_k+\mu u_k$ and $x_k-\mu u_k$ are evaluated using
samples from their respective induced distributions.

The two-point gradient estimator is
\[
    g_k
    =
    \frac{
    \ell(x_k+\mu u_k;z_k^+)
    -
    \ell(x_k-\mu u_k;z_k^-)
    }{2\mu}u_k,
\]
where
$z_k^+\sim D(x_k+\mu u_k)$ and
$z_k^-\sim D(x_k-\mu u_k)$.
The decision is then updated by
\[
    x_{k+1}=x_k-\eta_k g_k.
\]

Thus, guided exploration can improve efficiency when the
partial loss gradient aligns with the true performative-risk
gradient, but may become less effective when the two
are misaligned. Guided-ZO retains its
theoretical convergence guarantee without requiring
such alignment.

\paragraph{Discussion}
Consider the structural assumptions on the response distribution
$D(\cdot)$. The Two-stage algorithm assumes a specific location-family
response model. PerfGD allows a more general known parametric density
family $p(z;f(x))$; for example, the response distribution may be a
Gaussian mixture. Plug-in
assumes a specified parametric
family for the response map. DSA-PGM also assumes that
$D(x)$ belongs to a known parametric family with an unknown
decision-dependent parameter map.

In contrast, PTRO, DFO$(\lambda)$, and Guided-ZO do not assume that the response
distribution admits a fixed finite-dimensional parameterization. The response
distribution is treated as an unknown object that may vary with the
deployed decision, subject only to regularity conditions needed for the
analysis. 

Consider the theoretical guarantees. Two-stage and Plug-in
Performative Optimization target approximate performative optimality
and provide high-probability guarantees by exploiting additional
structure of the response model. PerfGD targets stationarity under
convexity and bounded response-gradient estimation error. DSA-PGM
allows a nonconvex performative risk and provides an expected
stationarity guarantee in terms of the projected-gradient mapping.
DFO$(\lambda)$ and Guided-ZO also allow nonconvex performative risks and establish stationarity guarantees in expectation. In comparison, PTRO establishes a high-probability $\epsilon$-stationarity guarantee for a general nonconvex performative risk without explicitly estimating or parameterizing $D(\cdot)$.
\subsection{Existing Response Models}
\addcontentsline{apc}{subsection}{\protect\numberline{\thesubsection}Existing Response Models}
\label{Existing_Response Models_as_Special_Cases}

We give some examples show that our settings recover several structured response models considered in prior work, including location-family, mixture-Gaussian, general parametric, and partially misspecified population responses. 
At the same time, our analysis does not require the learner to know or estimate these underlying response families: it only requires the corresponding moment or corruption conditions on the induced loss to hold.
\begin{example}[Mixture Gaussian response]
    Suppose $D(x) = \sum_{i=1}^K \gamma_i\,\mathcal N(\mu_i(x),\Sigma_i)$, $\sum_{i=1}^K\gamma_i=1$, $\gamma_i\ge0$, is a mixture of Gaussians with fixed covariances $\Sigma_i$ and means $\mu_i(x)$ that shift with the deployed decision. Consequently, for any loss $\ell$ with at most polynomial growth in $z$ (e.g., squared, logistic, or cross-entropy loss), $\ell(z;x)$ has a finite $q$-th moment for every $q\ge2$, so this setting falls under moment-bounded risk. 
\end{example}
This example includes the mixture-Gaussian and uniformly bounded loss assumptions used throughout~\citet{izzo2021learn}. It also includes the setting in~\citet{miller2021outside}: a location family $z_x \overset{d}{=} z_0+\mu x$ with sub-Gaussian $z_0$ .
\begin{example}[Non-Gaussian response]
    Suppose the response distribution belongs to a parametric family $\{D_{\beta_0}\}_{\beta_0\in B}$ (not necessarily Gaussian) whose members have uniformly upper-bounded finite moments up to order $q_{\max}$, i.e., $\sup_{\beta_0\in B}\mathbb E_{z\sim D_{\beta_0}}\|z\|^{q_{\max}}<\infty$. If the loss $\ell$ has at most polynomial growth in $z$, i.e., $|\ell(z;x)|\le K(1+\|z\|^{p})$ for some $p\le q_{\max}$, then $\ell(z;x)$ has a finite $q$-th moment for every $q\le q_{\max}/p$ and $L_q = K^q 2^{q-1}\Big(1+M_{q_{\max}}^{\,pq/q_{\max}}\Big)$, so this setting falls under moment-bounded risk. This includes the exact setting in~\citet{hamaguchi2026adaptive}. 
\end{example}

\begin{example}[Partially unmodeled response]
Consider
\[
    D(x)
    =
    (1-p)D_0(x)
    +
    p\,\widetilde D(x),
    \qquad p\in[0,1],
\]
where $D_0(x)$ describes a structured population response and
$\widetilde D(x)$ represents an unmodeled subpopulation.

This has the same mixture form as the ``typically well-specified''
population model of~\citet[Example~4]{lin2023plug}. In their plug-in framework, if the model atlas contains only the structured component $D_0(x)$,
then the remaining $p$-fraction is treated as model misspecification. Consequently, their guarantee contains an irreducible misspecification term that does not vanish with additional samples.

In contrast, PTRO does not attempt to represent $D(\cdot)$ through a response-model atlas. If the full mixture satisfies the required
loss-moment, stochastic-gradient, and decoupled-risk smoothness
conditions, then PTRO applies directly to the mixture distribution. Thus, the unmodeled component does not by itself introduce a separate
model-misspecification error in our analysis.
\end{example}
Therefore, our conditions include several
structured response models studied in prior work but not requiring
the learner to know their parametric form. 

\subsection{Relation to standard trust-region methods}
\addcontentsline{apc}{subsection}{\protect\numberline{\thesubsection}Relation to standard trust-region methods}
\label{appdix:relation_to_TR}

PTRO follows the standard stochastic trust-region architecture:
at each iteration, it constructs a local model, approximately minimizes the
model within a trust region, evaluates a trial point, and uses the agreement
between predicted and observed reduction to accept or reject the step and
adapt the trust-region radius. The main performative-specific modification is
the construction of the local model, which exploits the decoupled structure of
the performative risk.

\paragraph{Local approximation.}
A standard trust-region method constructs a local quadratic model of the full
objective,
\begin{equation}
    q_k(s)
    =
    PR(x_k)
    +
    g_k^\top s
    +
    \frac{1}{2}s^\top H_k s,
    \qquad
    \|s\| \le \alpha_k,
\end{equation}
where $g_k$ approximates $\nabla PR(x_k)$.

In performative optimization, the objective admits the decoupled representation
$PR(x)=R(x,x)$. At the current iterate,
\begin{align}
    PR(x_k+s)
    &=
    \underbrace{R(x_k+s,x_k)}_{A_k(s)}
    +
    \underbrace{
    \left[
    R(x_k+s,x_k+s)
    -
    R(x_k+s,x_k)
    \right]}_{\Phi_k(s)}.
\end{align}
The first component $A_k(s)$ can be estimated for multiple candidate steps
using the same samples from the current population $D(x_k)$, without additional
deployments. The second component captures the effect of the population
response and cannot in general be evaluated from samples at $D(x_k)$ alone.

Moreover, $\Phi_k(0)=0$ and
\begin{align}
    \nabla_s \Phi_k(0)
    &=
    \nabla_{x_1}R(x_k,x_k)
    +
    \nabla_{x_2}R(x_k,x_k)
    -
    \nabla_{x_1}R(x_k,x_k) \\
    &=
    \nabla_{x_2}R(x_k,x_k).
\end{align}
Hence,
\begin{equation}
    \Phi_k(s)
    =
    \nabla_{x_2}R(x_k,x_k)^\top s
    +
    O(\|s\|^2).
\end{equation}

This motivates the structured local model
\begin{equation}
    m_k(s)
    =
    \widehat R_k(x_k+s,x_k)
    +
    g_{2,k}^\top s
    +
    \frac{1}{2}s^\top H_k s,
    \qquad
    g_{2,k}
    \approx
    \nabla_{x_2}R(x_k,x_k),
\end{equation}
where $H_k$ is a bounded quadratic correction.

At the trust-region center,
\begin{align}
    \nabla_s m_k(0)
    &=
    \nabla_{x_1}\widehat R_k(x_k,x_k)+g_{2,k}\\
    &\approx
    \nabla_{x_1}R(x_k,x_k)
    +
    \nabla_{x_2}R(x_k,x_k)\\
    &=
    \nabla PR(x_k).
\end{align}
Thus, although the local model is constructed differently from the usual
quadratic approximation of the full objective, its gradient at the trust-region
center estimates the full performative gradient. Controlling the model-gradient
error therefore provides a direct connection between the trust-region model and
stationarity of the original performative risk.

\paragraph{Trust-region mechanism.}
Once the structured model is constructed, PTRO follows the usual stochastic
trust-region procedure. It computes a step satisfying a fraction-of-Cauchy-
decrease condition, evaluates the trial decision using deployment samples, and
accepts or rejects the step according to the agreement between predicted and
observed reduction. The radius is then enlarged or reduced using the standard
trust-region adaptation principle. The additional $2\epsilon_f$ correction in
the acceptance ratio accounts for stochastic error in the two performative-risk
estimates.

\paragraph{Performative oracle versus an abstract stochastic oracle.}
Standard stochastic trust-region analyses are often formulated in terms of
abstract model- and function-value estimators satisfying prescribed conditional
accuracy properties. In performative optimization, these properties are not
available directly: querying a decision changes the data distribution, and the
response-gradient estimator is constructed from samples collected under several
nearby deployments. In addition, samples from the current population may be
reused across the local model, gradient estimator, and acceptance test. Our
estimator analysis shows that this performative sampling procedure satisfies the
accuracy conditions needed for trust-region progress under the stated moment
and smoothness assumptions. Thus, the trust-region mechanism itself is standard;
the performative-specific contribution lies in constructing and analyzing a
structured stochastic oracle that exploits the known loss without specifying a
response model.

\paragraph{Step-acceptance mechanism}
The step-acceptance mechanism remains the same as in a standard trust-region method.
Given a standard local approximation $q_k$, the agreement between the realized and predicted
reductions is measured by
$$
    \rho_k
    =
    \frac{
        \widehat{PR}(x_k)
        -
        \widehat{PR}(x_k+s_k)
    }{
        q_k(0)-q_k(s_k)
    },
$$
where $\widehat{PR}$ denotes an observed estimate of the performative risk.
In our method, $m_k(\cdot)$ serves the same role as $q_k(\cdot)$: specifically,
$m_k(0)-m_k(s_k)$ represents the predicted reduction in the full performative risk.
We therefore retain the usual trust-region ratio and replace only the local approximation,
$$
    \rho_k
    =
    \frac{
        \widehat{PR}(x_k)
        -
        \widehat{PR}(x_k+s_k)
    }{
        m_k(0)-m_k(s_k)
    }.
$$
Thus, our variation modifies the construction of the local approximation and preserves
the standard trust-region principles for step generation and acceptance.

\section{Estimator Construction}
\addcontentsline{apc}{section}{\protect\numberline{\thesection}Estimator Construction}
\label{appdix:Estimator_Construction}
\subsection{Estimator accuracy guarantees and sample complexity}
\addcontentsline{apc}{subsection}{\protect\numberline{\thesubsection}Estimator accuracy guarantees and sample complexity}
\label{subsec:Estimator_accuracy_guarantees_sample_complexity}
For notational convenience, we write
$\mathrm{ZO}(\epsilon_f,q,\zeta_q)$ and
$\mathrm{FO}(\epsilon_{g,2},\delta_1)$
for the zeroth-order and first-order estimator conditions
in~\eqref{eq:moment-bdd-oracle} and~\eqref{eq:g_oracle},
respectively.

We now show that the estimators above satisfy the abstract
Estimator conditions required by PTRO.

\begin{lemma}[Sample complexity for ZO($\epsilon_f, q,\zeta_q$)]
Suppose the moment-bounded risk condition~\eqref{eq:moment_bdd_loss} holds. Given parameter $\epsilon_f > 0$,
 if the sample size satisfies $m_{\mathrm{ZO}} \ge \frac{L_q^{2/q}}{\epsilon_f^2} $, then $f_{m_{\mathrm{ZO}}}(x_1,x_2)$ is ZO$(\epsilon_f, q, \zeta_q)$, with $\zeta_q= 2^{q-1}\!\left(
        2^q C_q\, L_q^{\frac{2-q}{q}}\,\epsilon_f^{2(q-1)} 
        + (C_q+1)\,\epsilon_f^q
    \right)$ for some constant $C_q$.
\label{lemma:sample_cplx_zero_order_erre_bdd_moment}
\end{lemma}
In particular, the risk estimator can achieve arbitrarily small estimation error $\epsilon_f$ by increasing the minibatch size, with $m_{\mathrm{ZO}} =O(\epsilon_f^{-2}).$
\begin{proof}
For simplicity, we write $m$ for $m_{\mathrm{ZO}}$.

Condition on the algorithmic history available before the minibatch is
drawn. Given this history, the query pair $(x_1,x_2)$ is fixed and
$z_i\overset{i.i.d.}{\sim}D(x_2)$. Define
$Y_i:=\ell(z_i;x_1)-R(x_1,x_2),
\quad i=1,\dots,m.$

Conditionally on $\mathcal F_{k-1}$, the random variables
$Y_1,\dots,Y_m$ are i.i.d. and satisfy
\[
\mathbb E[Y_i\mid\mathcal F_{k-1}]=0.
\]
Moreover,
\[
e(x_1,x_2)
=
\left|
\frac{1}{m}\sum_{i=1}^m Y_i
\right|.
\]

We first bound the second and $q$-th moments of $Y_i$. By Jensen's inequality,
\[
|R(x_1,x_2)|^q
=
\left|
\mathbb E_{z\sim D(x_2)}[\ell(z;x_1)]
\right|^q
\le
\mathbb E_{z\sim D(x_2)}|\ell(z;x_1)|^q
\le L_q.
\]
By moment monotonicity,
\[
\mathbb E_{z\sim D(x_2)}|\ell(z;x_1)|^2
\le
\left(
\mathbb E_{z\sim D(x_2)}|\ell(z;x_1)|^q
\right)^{2/q}
\le
L_q^{2/q}.
\]

Since the $Y_i$ are conditionally independent and centered,
\begin{align*}
\mathbb E\left[
\left|
\frac{1}{m}\sum_{i=1}^m Y_i
\right|^2
\middle|
\mathcal F_{k-1}
\right]
&=
\frac{1}{m^2}
\sum_{i=1}^m
\mathbb E[
|Y_i|^2
\mid
\mathcal F_{k-1}
]
\\
&=
\frac{1}{m}
\mathbb E[
|Y_1|^2
\mid
\mathcal F_{k-1}
]
\\
&\le
\frac{
\mathbb E_{z\sim D(x_2)}
|\ell(z;x_1)|^2
}{m}
\le
\frac{L_q^{2/q}}{m}.
\end{align*}
Here,
\[
\mathbb E[
|Y_i|^2
\mid
\mathcal F_{k-1}
]
=
\operatorname{Var}_{z\sim D(x_2)}
(\ell(z;x_1))
\le
\mathbb E_{z\sim D(x_2)}
|\ell(z;x_1)|^2.
\]
Therefore,
\[
\mathbb E[
e
\mid
\mathcal F_{k-1}
]
\le
\left(
\mathbb E[
e^2
\mid
\mathcal F_{k-1}
]
\right)^{1/2}
\le
\frac{L_q^{1/q}}{\sqrt m}.
\]
Hence, if
\[
m\ge \frac{L_q^{2/q}}{\epsilon_f^2},
\]
then
\[
\mathbb E[
e
\mid
\mathcal F_{k-1}
]
\le \epsilon_f.
\]

Next, using
$|a-b|^q\le 2^{q-1}(|a|^q+|b|^q)$,
\begin{align*}
&\mathbb E\left[
\left|
e-\mathbb E[e\mid\mathcal F_{k-1}]
\right|^q
\middle|
\mathcal F_{k-1}
\right]
\\
&\qquad\le
2^{q-1}
\left(
\mathbb E[
e^q
\mid
\mathcal F_{k-1}
]
+
\left|
\mathbb E[
e
\mid
\mathcal F_{k-1}
]
\right|^q
\right).
\end{align*}

By Rosenthal's inequality~\cite{rosenthal1970subspaces}, there exists
a constant $C_q$ such that, for $q\ge2$,
\begin{align*}
\mathbb E[
e^q
\mid
\mathcal F_{k-1}
]
&=
\frac{1}{m^q}
\mathbb E\left[
\left|
\sum_{i=1}^m Y_i
\right|^q
\middle|
\mathcal F_{k-1}
\right]
\\
&\le
\frac{C_q}{m^q}
\left(
m\,
\mathbb E[
|Y_1|^q
\mid
\mathcal F_{k-1}
]
+
m^{q/2}
\left(
\mathbb E[
|Y_1|^2
\mid
\mathcal F_{k-1}
]
\right)^{q/2}
\right)
\\
&\le
C_q
\left(
\frac{2^qL_q}{m^{q-1}}
+
\frac{L_q}{m^{q/2}}
\right),
\end{align*}
where
\[
\mathbb E[
|Y_1|^q
\mid
\mathcal F_{k-1}
]
\le
2^{q-1}
\left(
\mathbb E_{z\sim D(x_2)}
|\ell(z;x_1)|^q
+
|R(x_1,x_2)|^q
\right)
\le
2^qL_q,
\]
and
\[
\mathbb E[
|Y_1|^2
\mid
\mathcal F_{k-1}
]
\le L_q^{2/q}.
\]

Under
$m\ge L_q^{2/q}/\epsilon_f^2$,
\[
\frac{L_q}{m^{q/2}}
\le \epsilon_f^q
\]
and
\[
\frac{L_q}{m^{q-1}}
\le
L_q^{\frac{2-q}{q}}
\epsilon_f^{2(q-1)}.
\]
Also,
\[
\left|
\mathbb E[
e
\mid
\mathcal F_{k-1}
]
\right|^q
\le
\epsilon_f^q.
\]
Therefore,
\begin{align*}
&\mathbb E\left[
\left|
e-\mathbb E[e\mid\mathcal F_{k-1}]
\right|^q
\middle|
\mathcal F_{k-1}
\right]
\\
&\qquad\le
2^{q-1}\left(
2^q C_q
L_q^{\frac{2-q}{q}}
\epsilon_f^{2(q-1)}
+
(C_q+1)\epsilon_f^q
\right)
=:\zeta_q.
\end{align*}
\end{proof}

\begin{lemma}[Parameter for FO]
\label{lemma:FO_construction}
\added{Under Assumption~\ref{assp:nabla_x_2_R_lip}} and the uniform
$q$-th moment bound with $q\ge 2$,
construct $G_2$ using coordinate-wise
finite differences with minibatch size
$m_{\mathrm{ZO}}$.

For any $\epsilon_{g,2}>0$ and
$\delta_1\in(0,1)$, set
$
\sigma =
\sqrt{\frac{\delta_1\epsilon_{g,2}^2}
{dL_{22}^2}}.
$ 
If
$
m_{\mathrm{ZO}}
\ge
\frac{8d^2L_{22}^2L_q^{2/q}}
{\delta_1^2\epsilon_{g,2}^4},
$
then
\[
\mathbb P\left(
\|G_2(x_1,x_2)
-\nabla_{x_2}R(x_1,x_2)\|
\le\epsilon_{g,2}
\mid\mathcal F_{k-1}
\right)
\ge 1-\delta_1.
\]
\end{lemma}


\begin{proof}
Suppose $(x_1,x_2)$ is $\mathcal F_{k-1}$-measurable. Let $\partial_i R(x_1,x_2)
:=
[\nabla_{x_2}R(x_1,x_2)]_i.$

For each coordinate $i$,
\begin{align*}
\left|[G_2]_i-\partial_i R(x_1,x_2)\right|
&\le
\left|
\frac{R(x_1,x_2+\sigma e_i)-R(x_1,x_2)}{\sigma}
-\partial_iR(x_1,x_2)
\right| \\
&\qquad
+
\frac{
e(x_1,x_2+\sigma e_i)+e(x_1,x_2)
}{\sigma}\\
&\le
\frac{L_{22}}{2}\sigma
+
\frac{
e(x_1,x_2+\sigma e_i)+e(x_1,x_2)
}{\sigma}.
\end{align*}

\added{Let}
\[
\widehat R_0=f_{m_{\mathrm{ZO}}}(x_1,x_2),
\qquad
\widehat R_i=f_{m_{\mathrm{ZO}}}(x_1,x_2+\sigma e_i),
\]
and define the centered sampling errors
\[
\xi_0=\widehat R_0-R(x_1,x_2),\qquad
\xi_i=\widehat R_i-R(x_1,x_2+\sigma e_i).
\]
Since $q\ge 2$, the uniform moment condition implies
\[
\mathbb E[\xi_i^2\mid\mathcal F_{k-1}]
\le \frac{L_q^{2/q}}{m_{\mathrm{ZO}}},
\qquad i=0,\ldots,d.
\]

Decompose the gradient estimation error as
\[
G_2-\nabla_{x_2}R=B_\sigma+S_\sigma,
\]
where $B_\sigma$ is the finite-difference bias and
$
[S_\sigma]_i=\frac{\xi_i-\xi_0}{\sigma}.
$
By Assumption~\ref{assp:nabla_x_2_R_lip},
\[
\|B_\sigma\|^2\le\frac{dL_{22}^2\sigma^2}{4}.
\]
Since the base and perturbed minibatches are
conditionally independent,
\[
\begin{aligned}
\mathbb E[\|S_\sigma\|^2\mid\mathcal F_{k-1}]
&=\frac{1}{\sigma^2}
\sum_{i=1}^d
\mathbb E[(\xi_i-\xi_0)^2\mid\mathcal F_{k-1}]\\
&\le \frac{2dL_q^{2/q}}{m_{\mathrm{ZO}}\sigma^2}.
\end{aligned}
\]
Consequently,
$$
\mathbb E[
\|G_2-\nabla_{x_2}R\|^2
\mid\mathcal F_{k-1}]
\quad\le
\frac{dL_{22}^2\sigma^2}{2}
+\frac{4dL_q^{2/q}}{m_{\mathrm{ZO}}\sigma^2}.
$$

By conditional Markov's inequality,
$$
\mathbb P\left(
\|G_2-\nabla_{x_2}R\|>\epsilon_{g,2}
\mid\mathcal F_{k-1}
\right)
\quad\le
\frac{1}{\epsilon_{g,2}^2}
\left(
\frac{dL_{22}^2\sigma^2}{2}
+\frac{4dL_q^{2/q}}{m_{\mathrm{ZO}}\sigma^2}
\right).
$$

Thus, it suffices to choose $\sigma$ and
$m_{\mathrm{ZO}}$ such that
\[
\frac{dL_{22}^2\sigma^2}{2}
+\frac{4dL_q^{2/q}}{m_{\mathrm{ZO}}\sigma^2}
\le\delta_1\epsilon_{g,2}^2.
\]
Under this condition,
$
\mathbb P\left(
\|G_2-\nabla_{x_2}R(x_1,x_2)\|
\le\epsilon_{g,2}
\mid\mathcal F_{k-1}
\right)\ge1-\delta_1.
$

\end{proof}

\subsection{Practical Conditions for Estimator Accuracy}
\addcontentsline{apc}{subsection}{\protect\numberline{\thesubsection}Practical Conditions for Estimator Accuracy}
\label{subsec:Practical Conditions for Estimator Accuracy}
Here, we discuss two assumptions used to control the estimator errors
and their practical interpretation.

Assumption~\ref{assp:gradient_var_bdd} controls the sampling
variability of the direct gradient.
This condition does not require a parametric model for $D(x_2)$.
A simple sufficient condition is a uniformly bounded sample gradient,
$\|\nabla_{x_1}\ell(z;x_1)\|\le G$, which implies the variance is at
most $G^2$. For example, under logistic loss,
\[
\ell(z;x)=\log\!\left(1+\exp(-y x^\top a)\right),
\]
we have $\|\nabla_x\ell(z;x)\|\le \|a\|$. Hence, bounded or clipped
features directly support this assumption.

Assumption~\ref{assp:nabla_x_2_R_lip} requires
$\nabla_{x_2}R(x_1,x_2)$ to vary Lipschitz-continuously with the
deployment $x_2$. Intuitively, small changes in deployment should induce
smooth changes in the loss-relevant population statistics. This can be
encouraged by using losses that depend smoothly on the observed data.
For instance, if
\[
\ell(z;x_1)=a(x_1)^\top\phi(z)+b(x_1),
\]
then
\[
R(x_1,x_2)
=
a(x_1)^\top\mu_\phi(x_2)+b(x_1),
\qquad
\mu_\phi(x_2)
=
\mathbb E_{z\sim D(x_2)}[\phi(z)].
\]
Thus, a Lipschitz Jacobian of $\mu_\phi(\cdot)$ is sufficient for
Lipschitz continuity of $\nabla_{x_2}R$, without specifying a
parametric form for $D(x_2)$.

In practice, smooth losses and bounded or normalized features are
therefore preferable. Logistic loss is a natural choice for
classification, smooth pairwise logistic or softmax losses
are similarly suitable for ranking or recommendation problems.

\section{Proofs for Convergence}
\label{appdix:proof_for_convergence}
\addcontentsline{apc}{section}{\protect\numberline{\thesection}Proofs for Convergence and Complexity}

\subsection{Proof of Lemma~\ref{lemma:p_lower_bd}}
\addcontentsline{apc}{subsection}
{\protect\numberline{\thesubsection}Proof of Lemma~\ref{lemma:p_lower_bd}}
\begin{proof}
Define the accuracy events
\begin{align*}
\mathcal E_{k,1}
&:=
\left\{
\|\nabla_{x_1}f_{m_{\mathrm{ZO}}}(X_k,X_k)
-\nabla_{x_1}R(X_k,X_k)\|
\le\epsilon_{g,1}
\right\},\\
\mathcal E_{k,2}
&:=
\left\{
\|g_{2,k}-\nabla_{x_2}R(X_k,X_k)\|
\le\epsilon_{g,2}
\right\},\\
\mathcal E_{k,f}
&:=\{e_k+e_k^+\le2\epsilon_f\}.
\end{align*}
By definition,
\[
\{I_k=1\}
=\mathcal E_{k,1}\cap\mathcal E_{k,2}
 \cap\mathcal E_{k,f}.
\]

Conditional on $\mathcal F_{k-1}$, the current point
$X_k$ is fixed and the base samples are independent
with distribution $D(X_k)$.
The unbiasedness of the direct-gradient estimator
and Assumption~4 therefore give
\[
\mathbb E\!\left[
\|\nabla_{x_1}f_{m_{\mathrm{ZO}}}(X_k,X_k)
-\nabla_{x_1}R(X_k,X_k)\|^2
\,\middle|\,\mathcal F_{k-1}
\right]
\le \frac{\nu^2}{m_{\mathrm{ZO}}}.
\]
Hence, by Markov's inequality,
\[
\mathbb P(\mathcal E_{k,1}^c
\mid\mathcal F_{k-1})
\le
\frac{\nu^2}{m_{\mathrm{ZO}}\epsilon_{g,1}^2}.
\]
The assumed response-gradient accuracy gives
\[
\mathbb P(\mathcal E_{k,2}^c
\mid\mathcal F_{k-1})\le\delta_1.
\]

For the base risk estimate, the moment condition and
the variance bound for a sample mean imply
\[
\mathbb E[e_k^2\mid\mathcal F_{k-1}]
\le \frac{L_q^{2/q}}{m_{\mathrm{ZO}}}.
\]

To handle the trial estimate, let $\mathcal G_k$
denote the history after constructing $X_k^+$ but
before drawing its evaluation samples.
Then $\mathcal F_{k-1}\subseteq\mathcal G_k$ and
$X_k^+$ is $\mathcal G_k$-measurable.
Conditional on $\mathcal G_k$, the trial samples
are independent with distribution $D(X_k^+)$.
The uniform moment condition therefore gives
\[
\mathbb E[(e_k^+)^2\mid\mathcal G_k]
\le \frac{L_q^{2/q}}{m_{\mathrm{ZO}}}.
\]
By the tower property,
\[
\mathbb E[(e_k^+)^2\mid\mathcal F_{k-1}]
\le \frac{L_q^{2/q}}{m_{\mathrm{ZO}}}.
\]

Since
\[
\{e_k+e_k^+>2\epsilon_f\}
\subseteq
\{e_k>\epsilon_f\}\cup\{e_k^+>\epsilon_f\},
\]
Markov's inequality and the union bound yield
\[
\mathbb P(\mathcal E_{k,f}^c
\mid\mathcal F_{k-1})
\le
\frac{2L_q^{2/q}}{m_{\mathrm{ZO}}\epsilon_f^2}.
\]
Finally, applying the union bound to the three
accuracy events gives
\begin{align*}
\mathbb P(I_k=1\mid\mathcal F_{k-1})
&\ge
1-\mathbb P(\mathcal E_{k,1}^c\mid\mathcal F_{k-1})
-\mathbb P(\mathcal E_{k,2}^c\mid\mathcal F_{k-1})
-\mathbb P(\mathcal E_{k,f}^c\mid\mathcal F_{k-1})\\
&\ge
1-\frac{\nu^2}{m_{\mathrm{ZO}}\epsilon_{g,1}^2}
-\delta_1
-\frac{2L_q^{2/q}}{m_{\mathrm{ZO}}\epsilon_f^2}
\end{align*}
No independence between the three accuracy events
is required.
\end{proof}

\subsection{Model Accuracy and progress Guarantees}
\addcontentsline{apc}{subsection}
{\protect\numberline{\thesubsection}Model Accuracy and progress Guarantees}
\paragraph{Notation and radius threshold.}
Let
\[
    v_k:=\nabla m_k(0),\qquad
    M:=L_\ell+\kappa_H,\qquad
    c:=\frac{\kappa_{\mathrm{fcd}}}{2}.
\]
For a step $s$, define the predicted and actual risk reductions by
\[
    \Delta m_k(s):=m_k(0)-m_k(s),\qquad
    \Delta PR_k(s):=PR(x_k)-PR(x_k+s).
\]
Write
\[
    \epsilon_g:=\epsilon_{g,1}+\epsilon_{g,2},
\]
For a target tolerance $\epsilon>0$, define
\[
    T_\epsilon:=
    \inf\{k\geq0:\|\nabla PR(x_k)\|\leq\epsilon\},
\]
Thus, $\|\nabla PR(x_k)\|>\epsilon$ whenever $k<T_\epsilon$.

Define the radius threshold
\[
    \bar{\alpha}:=C_\alpha\epsilon,
\]
where
\begin{equation}
    C_\alpha:=
    \min\left\{
        \frac{1-\eta_g}{M},
        \frac{2\bigl[c(1-\eta_1)(1-\eta_g)-\eta_g\bigr]}{L+M},
        \frac{1-\eta_g}{\eta_2}
    \right\}.
    \label{eq:radius_threshold_fcd}
\end{equation}

\begin{definition}[Large and small steps]
For the threshold $\bar{\alpha} > 0$, we define \textbf{small step} when $\max\{\alpha_k,\alpha_{k+1}\} \le \bar{\alpha}$ otherwise is \textbf{large step}.
\end{definition}

\begin{lemma}[Model-reduction accuracy]
\label{lemma:model_reduction_accuracy}
   Under Assumption~\ref{assp:PR_smooth},~\ref{assp:H_k_bound}, and~\ref{assp:Smoothness_in_loss_function} , if $k$-th iteration is true, then:
   $$| \Delta PR_k(s) - \Delta m_k(s)| \le \frac{M + L}{2}\|s\|^2 + \epsilon_g \|s\|$$
\end{lemma}
\begin{proof}

By the definition of a true iteration and the identity
$\nabla PR(x_k)=
\nabla_{x_1}R(x_k,x_k)+\nabla_{x_2}R(x_k,x_k)$. Recall $v_k=\nabla m_k(0)=\nabla_{x_1}\widehat R_k(x_k,x_k)+g_{2,k},$ on a true iteration,
\[
    \|v_k-\nabla PR(x_k)\|\leq\epsilon_g.
\]
Since the empirical loss is $L_\ell$-smooth and
$\|H_k\|\leq\kappa_H$, the model $m_k$ is $M$-smooth.

By $L$-smoothness of $PR$,
\[
\left|
PR(x_k+s)-PR(x_k)-\nabla PR(x_k)^\top s
\right|
\le \frac{L}{2}\|s\|^2.
\]
Similarly, since $m_k$ is $M$-smooth,
\[
\left|
m_k(s)-m_k(0)-v_k^\top s
\right|
\le \frac{M}{2}\|s\|^2.
\]
Therefore,
\[
\begin{aligned}
|\Delta PR_k(s)-\Delta m_k(s)|
&\le
\|\nabla PR(x_k)-v_k\|\,\|s\|
+\frac{L+M}{2}\|s\|^2 \\
&\le
\epsilon_g\|s\|
+\frac{L+M}{2}\|s\|^2.
\end{aligned}
\]

\end{proof}

\begin{definition}[Successful iteration]
\label{def:successful_iteration}
Let $A_k$ denote the random trust-region radius at iteration $k$,
with realization $\alpha_k$.
An iteration $k$ is called successful if the trial point is accepted and
the trust-region radius is increased, that is,
\[
X_{k+1}=X_k^+
\qquad\text{and}\qquad
A_{k+1}=\gamma_{\mathrm{inc}}A_k.
\]
Equivalently, for PTRO this occurs when
$\rho_k\ge \eta_1
\quad\text{and}\quad
\|V_k\|\ge \eta_2 A_k,$

where $V_k:=\nabla m_k(0)$ denotes the random model gradient.
\end{definition}

\begin{lemma}[small true iterations are successful.]
    Under Assumption~\ref{assp:H_k_bound},~\ref{assp:PR_smooth}, ,and  \ref{assp:Smoothness_in_loss_function}, any true iteration $k < T_\epsilon$, If $\alpha_k\leq\bar\alpha$, then 
$$
    \rho_k\geq\eta_1,
    \quad \|v_k\|\geq\eta_2\alpha_k.
$$
Consequently, the iteration is successful.
\label{lemma:small_true_must_successful}
\end{lemma}
\begin{proof}
Recall $\epsilon_g \le \eta_g \epsilon.$ Since $k<T_\epsilon$ and the iteration is true,
$$
    \|v_k\|
    \geq
    \|\nabla PR(x_k)\|-\epsilon_g
    >
    (1-\eta_g)\epsilon.
$$
The first term in~\eqref{eq:radius_threshold_fcd} implies, when $\alpha_k \le \bar \alpha$
$$
    \alpha_k\leq
    \frac{(1-\eta_g)\epsilon}{M}
    \leq\frac{\|v_k\|}{M}.
$$
Thus, the Cauchy-decrease condition gives
\begin{equation}
    \Delta m_k(s_k)
    \geq
    c\|v_k\|
    \min\left\{\frac{\|v_k\|}{M},\alpha_k\right\}
    = c\|v_k\|\alpha_k
    \geq c(1-\eta_g)\epsilon\alpha_k>0.
    \label{eq:small_predicted_decrease}
\end{equation}

On a true iteration, $e_k+e_k^+\leq2\epsilon_f$, so
\[
    f_k-f_k^++2\epsilon_f
    \geq\Delta PR_k(s_k).
\]
Using Lemma~\ref{lemma:model_reduction_accuracy},
$\|s_k\|\leq\alpha_k$, and
\eqref{eq:small_predicted_decrease}, we have
\[
\begin{aligned}
    \rho_k
    &\geq
    \frac{\Delta PR_k(s_k)}{\Delta m_k(s_k)}
    \\
    &\geq
    1-
    \frac{\epsilon_g\alpha_k+\frac{L+M}{2}\alpha_k^2}
         {c(1-\eta_g)\epsilon\alpha_k}
    \\
    &\geq
    1-
    \frac{\eta_g+\frac{L+M}{2}\alpha_k/\epsilon}
         {c(1-\eta_g)}
    \geq\eta_1.
\end{aligned}
\]
The final inequality follows from the second term in
\eqref{eq:radius_threshold_fcd}.

Finally, the third term gives
\[
    \eta_2\alpha_k
    \leq(1-\eta_g)\epsilon
    \leq\|v_k\|.
\]
Both conditions for a successful iteration therefore hold.
\end{proof}

\begin{lemma}[Progress on large successful iterations]
\label{lemma:h(eps)}
Let $M := L_\ell+\kappa_H$ and
$\bar{\alpha}:=C_\alpha\epsilon$, where $C_\alpha>0$.
Define
\[
h(\epsilon):=C_h\epsilon^2,
\qquad
C_h:=
\frac{\eta_1\kappa_{\mathrm{fcd}}\eta_2}{2}
\min\left\{\frac{\eta_2}{M},1\right\}
\left(\frac{C_\alpha}{\gamma_{\mathrm{inc}}}\right)^2
>0.
\]
Suppose the step $s_k$ satisfies the fraction-of-Cauchy-decrease
condition. If iteration $k$ is successful and large, then
\[
PR(x_k)-PR(x_{k+1})
\ge h(\epsilon)-(2\epsilon_f+e_k+e_k^+).
\]
Here, successful means that
$\rho_k\ge\eta_1$ and $\|v_k\|\ge\eta_2\alpha_k$,
where $v_k:=\nabla m_k(0)$.
\end{lemma}

\begin{proof}
Since iteration $k$ is successful,
$x_{k+1}=x_k+s_k$,
$\alpha_{k+1}=\gamma_{\mathrm{inc}}\alpha_k$,
and $\|v_k\|\ge\eta_2\alpha_k$.
Because the iteration is also large,
\[
\max\{\alpha_k,\alpha_{k+1}\}>\bar{\alpha}.
\]
Hence $\alpha_k>\bar{\alpha}/\gamma_{\mathrm{inc}}$.

By the fraction-of-Cauchy-decrease condition,
\begin{align*}
\Delta m_k
&:=m_k(0)-m_k(s_k)\\
&\ge
\frac{\kappa_{\mathrm{fcd}}}{2}\|v_k\|
\min\left\{\frac{\|v_k\|}{M},\alpha_k\right\}\\
&\ge
\frac{\kappa_{\mathrm{fcd}}\eta_2}{2}
\min\left\{\frac{\eta_2}{M},1\right\}\alpha_k^2\\
&\ge
\frac{\kappa_{\mathrm{fcd}}\eta_2}{2}
\min\left\{\frac{\eta_2}{M},1\right\}
\left(\frac{\bar{\alpha}}{\gamma_{\mathrm{inc}}}\right)^2
>0.
\end{align*}

Write $f_k$ and $f_k^+$ for the risk estimates at $x_k$
and $x_k+s_k$, respectively. The acceptance condition gives
\[
f_k-f_k^++2\epsilon_f
\ge \eta_1\Delta m_k.
\]
By the definitions of $e_k$ and $e_k^+$,
\[
f_k-f_k^+
\le PR(x_k)-PR(x_{k+1})+e_k+e_k^+.
\]
Combining these inequalities yields
\begin{align*}
PR(x_k)-PR(x_{k+1})
&\ge \eta_1\Delta m_k-(2\epsilon_f+e_k+e_k^+)\\
&\ge
\frac{\eta_1\kappa_{\mathrm{fcd}}\eta_2}{2}
\min\left\{\frac{\eta_2}{M},1\right\}
\left(\frac{\bar{\alpha}}{\gamma_{\mathrm{inc}}}\right)^2
-(2\epsilon_f+e_k+e_k^+)\\
&=h(\epsilon)-(2\epsilon_f+e_k+e_k^+),
\end{align*}
where the last equality uses $\bar{\alpha}=C_\alpha\epsilon$.
\end{proof}

\subsection{Proof of Proposition \ref{prop:lower_bd_for_eps}}
\addcontentsline{apc}{subsection}{\protect\numberline{\thesubsection}Proof of Proposition \ref{prop:lower_bd_for_eps}}
\label{appdix:proof_for_prop_Attainable_eps}
\begin{proof}
Let $\epsilon>0$ be the target stationarity tolerance, and write
\[
\epsilon_g=\epsilon_{g,1}+\epsilon_{g,2},
\qquad h(\epsilon)=C_h\epsilon^2,
\qquad
u=-\frac{\ln\gamma_{\mathrm{inc}}}{\ln\gamma_{\mathrm{dec}}}>0,
\]
where $C_h>0$ is the progress constant from the convergence analysis.

We first explain the threshold $p_u$. Before reaching an $\epsilon$-stationary point, the iteration-counting argument gives
\[
\sum_{k=0}^{t-1}I_k
\le
\frac{t}{u+1}
+\frac{u}{h(\epsilon)}
\left[
Z_0+\sum_{k=0}^{t-1}
\bigl(2\epsilon_f+e_k+e_k^+\bigr)
\right]
+D_\alpha,
\]
where $I_k$ indicates a true iteration, $Z_0=PR(X_0)-PR_*$, and $D_\alpha$ is the initial-radius term.

The quantity
$2\epsilon_f+E_k+E_k^+$
bounds the loss of progress caused by the relaxed acceptance test and
the two function-estimation errors. By the conditional ZO bounds,
\[
\mathbb E[E_k+E_k^+\mid\mathcal F_{k-1}]
\le 2\epsilon_f
\qquad\text{a.s.}
\]
Define
\[
Y_k
:=
(E_k+E_k^+)
-
\mathbb E[E_k+E_k^+\mid\mathcal F_{k-1}],
\qquad
S_t:=\sum_{k=0}^{t-1}Y_k.
\]
Then
\[
2\epsilon_f+E_k+E_k^+
=
2\epsilon_f
+
\mathbb E[E_k+E_k^+\mid\mathcal F_{k-1}]
+
Y_k
\le
4\epsilon_f+Y_k.
\]
Therefore,
\[
\sum_{k=0}^{t-1}
(2\epsilon_f+E_k+E_k^+)
\le
4\epsilon_f t+S_t,
\]
and the counting inequality gives
\[
\sum_{k=0}^{t-1}I_k
\le
\underbrace{\left(
\frac{1}{u+1}
+\frac{4u\epsilon_f}{h(\epsilon)}
\right)}_{p_u}t
+\frac{uZ_0}{h(\epsilon)}
+D_\alpha
+\frac{u}{h(\epsilon)}S_t.
\]
Thus,
\[
p_u:=\frac{1}{u+1}+\frac{4u\epsilon_f}{h(\epsilon)}
\]
is the baseline true-iteration frequency that must be exceeded to accumulate progress. The convergence analysis controls the fluctuations of $W_t$ and the
number of true iterations; the condition $p>p_u$ provides the required
positive margin.

We now verify the conditions in the proposition. By assumption,
$\epsilon_g<\eta_g\epsilon,$

so the gradient-estimation error is sufficiently small relative to the target tolerance. The restriction
\[
0<\eta_g<
\frac{\kappa_{\mathrm{fcd}}(1-\eta_1)}
     {2+\kappa_{\mathrm{fcd}}(1-\eta_1)}
\]
ensures that the radius-threshold constant $C_\alpha$ is positive.

Moreover, because $p>1/(u+1)$, the assumed bound
\[
\epsilon_f<
\frac{C_h}{4u}
\left(p-\frac{1}{u+1}\right)\epsilon^2
\]
has a positive right-hand side. Dividing by
$h(\epsilon)=C_h\epsilon^2>0$ and multiplying by $4u$ yields
\[
\frac{4u\epsilon_f}{h(\epsilon)}
<
p-\frac{1}{u+1}.
\]
Consequently,
\[
p_u
=\frac{1}{u+1}
+\frac{4u\epsilon_f}{h(\epsilon)}
<p.
\]
Both required conditions, $\epsilon_g<\eta_g\epsilon$ and $p>p_u$, therefore hold, so $\epsilon$ is admissible for the convergence guarantee.   
\end{proof} 

\subsection{Proof of Theorem \ref{thm:convergence}}
\addcontentsline{apc}{subsection}{
\protect\numberline{\thesubsection}
Proof of Theorem \ref{thm:convergence}}
\label{proof_for_thm_converge}

\begin{proof}
Lemma~\ref{lemma:true_iteration_probability} gives
\[
\mathbb P(I_k=1\mid\mathcal F_{k-1})\ge p
\]
almost surely on $\{k<T_\epsilon\}$.
Lemma~\ref{lemma:small_true_must_successful} shows that every small
true iteration before $T_\epsilon$ is successful.
Lemma~\ref{lemma:h(eps)} shows that, on every large successful
iteration,
\[
PR(X_k)-PR(X_{k+1})
\ge
h(\epsilon)-(2\epsilon_f+E_k+E_k^+),
\qquad
h(\epsilon)=C_h\epsilon^2.
\]
Moreover, on every iteration,
\[
PR(X_k)-PR(X_{k+1})
\ge
-(2\epsilon_f+E_k+E_k^+).
\]

The admissibility conditions give
\[
\epsilon_g<\eta_g\epsilon,
\qquad
p>p_u,
\qquad
p_u:=
\frac{1}{u+1}
+\frac{4u\epsilon_f}{h(\epsilon)}.
\]

We first control the accumulated function-estimation error.
Define
\[
Y_k
:=
(E_k+E_k^+)
-
\mathbb E[E_k+E_k^+\mid\mathcal F_{k-1}],
\qquad
S_t:=\sum_{k=0}^{t-1}Y_k.
\]
By construction,
\[
\mathbb E[Y_k\mid\mathcal F_{k-1}]=0,
\]
so $(Y_k,\mathcal F_k)$ is a martingale-difference sequence.
By the conditional ZO bounds,
\[
\mathbb E[|Y_k|^q\mid\mathcal F_{k-1}]
\le 2^q\zeta_q
\qquad\text{a.s.}
\]
Let $
\zeta_2:=\frac{L_q^{2/q}}{m_{\mathrm{ZO}}}.
$
By the second-moment bound established in
Lemma~\ref{lemma:sample_cplx_zero_order_erre_bdd_moment},
\[
\mathbb E[E_k^2\mid\mathcal F_{k-1}]
\le \zeta_2,
\qquad
\mathbb E[(E_k^+)^2\mid\mathcal F_{k-1}]
\le \zeta_2.
\]
Since
\[
Y_k
=
(E_k+E_k^+)
-
\mathbb E[E_k+E_k^+\mid\mathcal F_{k-1}],
\]
we have
\begin{align*}
\mathbb E[Y_k^2\mid\mathcal F_{k-1}]
&=
\operatorname{Var}
\bigl(E_k+E_k^+\mid\mathcal F_{k-1}\bigr)\\
&\le
\mathbb E\!\left[
(E_k+E_k^+)^2
\mid\mathcal F_{k-1}
\right]\\
&\le
2\mathbb E[E_k^2\mid\mathcal F_{k-1}]
+
2\mathbb E[(E_k^+)^2\mid\mathcal F_{k-1}]\\
&\le 4\zeta_2.
\end{align*}

For $s>0$, define the event
\[
\mathcal B_t(s)
:=
\left\{
\sum_{k=0}^{t-1}
(2\epsilon_f+E_k+E_k^+)
\le
(4\epsilon_f+s)t
\right\}.
\]
Since
\[
\mathbb E[E_k+E_k^+\mid\mathcal F_{k-1}]
\le 2\epsilon_f,
\]
we have
\[
\sum_{k=0}^{t-1}
(2\epsilon_f+E_k+E_k^+)
\le
4\epsilon_f t+S_t.
\]
Hence,
\[
\mathcal B_t(s)^c
\subseteq
\{S_t>st\}.
\]
Applying the martingale Fuk--Nagaev inequality yields
\[
\mathbb P\bigl(\mathcal B_t(s)^c\bigr)
\le
\delta_t(s),
\]
where
\[
\delta_t(s)
:=
\exp\!\left(
-\frac{s^2}{2(q+2)^2e^q\zeta_2}\,t
\right)
+
\left(1+\frac{2}{q}\right)^q
\frac{2^q\zeta_q}{s^q t^{q-1}}.
\]

We next control the number of true iterations.
For any
\[
\widehat p
\in
\left(
p_u+\frac{us}{h(\epsilon)},\,p
\right),
\]
define
\[
\mathcal A_t
:=
\left\{
\sum_{k=0}^{t-1}I_k
\ge
\widehat p\,t
\right\}.
\]
Consider the stopped process
\[
M_\ell
:=
\sum_{k=0}^{\ell-1}
\mathbf 1_{\{k<T_\epsilon\}}(I_k-p).
\]
Since $\{k<T_\epsilon\}\in\mathcal F_{k-1}$ and
\[
\mathbb E[I_k-p\mid\mathcal F_{k-1}]
\ge0
\quad\text{on }\{k<T_\epsilon\},
\]
$(M_\ell)$ is a submartingale.

Moreover, before $T_\epsilon$,
\[
I_k-p\in\{-p,1-p\},
\]
so
\[
|I_k-p|
\le
\max\{p,1-p\}.
\]
Therefore, Azuma--Hoeffding gives
\[
\mathbb P\!\left(
T_\epsilon>t,\,
\mathcal A_t^c
\right)
\le
\exp\!\left(
-\frac{(p-\widehat p)^2}
{2\max\{p,1-p\}^2}\,t
\right).
\]

Now suppose that
$T_\epsilon>t$
and that both $\mathcal A_t$ and $\mathcal B_t(s)$ occur.
By the iteration-counting argument,
\[
\sum_{k=0}^{t-1}I_k
\le
\frac{t}{u+1}
+
\frac{u}{h(\epsilon)}
\left[
Z_0+
\sum_{k=0}^{t-1}
(2\epsilon_f+E_k+E_k^+)
\right]
+
D_\alpha,
\]
where
\[
D_\alpha
:=
\frac{u}{u+1}
\max\left\{
0,\,
\left\lceil
\frac{\ln(\alpha_0/\bar\alpha)}
{\ln(1/\gamma_{\mathrm{dec}})}
\right\rceil
\right\}.
\]

On $\mathcal B_t(s)$,
\[
\sum_{k=0}^{t-1}
(2\epsilon_f+E_k+E_k^+)
\le
(4\epsilon_f+s)t.
\]
Hence,
\begin{align*}
\sum_{k=0}^{t-1}I_k
&\le
\frac{uZ_0}{h(\epsilon)}
+D_\alpha
+
\left[
\frac{1}{u+1}
+
\frac{u(4\epsilon_f+s)}{h(\epsilon)}
\right]t\\
&=
B_\epsilon
+
\left(
p_u+\frac{us}{h(\epsilon)}
\right)t,
\end{align*}
where
\[
B_\epsilon
:=
\frac{uZ_0}{h(\epsilon)}
+
\frac{u}{u+1}
\max\left\{
0,\,
\left\lceil
\frac{\ln(\alpha_0/\bar\alpha)}
{\ln(1/\gamma_{\mathrm{dec}})}
\right\rceil
\right\}.
\]

On the other hand, $\mathcal A_t$ implies
\[
\sum_{k=0}^{t-1}I_k
\ge
\widehat p\,t.
\]
Thus,
\[
\widehat p\,t
\le
B_\epsilon
+
\left(
p_u+\frac{us}{h(\epsilon)}
\right)t,
\]
or equivalently,
\[
t
\le
\frac{B_\epsilon}
{\widehat p-p_u-us/h(\epsilon)}.
\]
This contradicts the assumed condition
\[
t>
\frac{B_\epsilon}
{\widehat p-p_u-us/h(\epsilon)}.
\]
Therefore,
\[
\mathbb P(
T_\epsilon>t,\,
\mathcal A_t,\,
\mathcal B_t(s)
)=0.
\]

Finally,
\begin{align*}
\mathbb P(T_\epsilon>t)
&\le
\mathbb P(
T_\epsilon>t,\mathcal A_t^c)
+
\mathbb P(\mathcal B_t(s)^c)\\
&\le
\exp\!\left(
-\frac{(p-\widehat p)^2}
{2\max\{p,1-p\}^2}\,t
\right)
+
\delta_t(s).
\end{align*}
Hence,
\[
\mathbb P(T_\epsilon\le t)
\ge
1-
\exp\!\left(
-\frac{(p-\widehat p)^2}
{2\max\{p,1-p\}^2}\,t
\right)
-
\delta_t(s),
\]

\end{proof}

\subsection{Parameters Selection}
\addcontentsline{apc}{subsection}{\protect\numberline{\thesubsection}Parameters Selection}

\label{appdx:alg_parameters_selection}

We choose the estimator parameters for a 
stationarity tolerance $\epsilon\in(0,1)$.
Fix an error-allocation constant $\theta\in(0,1)$ and
a target lower bound on the probability of a true iteration
\[
p_0\in\left(\frac{1}{u+1},1\right),
\qquad
\Delta_p:=p_0-\frac{1}{u+1}>0.
\]
Our goal is to ensure
\[
\epsilon_g<\eta_g\epsilon,
\qquad
p\ge p_0,
\qquad
\frac{4u\epsilon_f}{h(\epsilon)}
\le\frac{\Delta_p}{2}.
\]
These conditions imply
\[
p_u
=\frac{1}{u+1}+\frac{4u\epsilon_f}{h(\epsilon)}
\le\frac{1}{u+1}+\frac{\Delta_p}{2}
<p_0\le p.
\]

\paragraph{Estimator accuracies.}
Set
\[
\epsilon_{g,1}:=\theta\eta_g\epsilon,
\qquad
\epsilon_{g,2}:=\frac{1-\theta}{2}\eta_g\epsilon,
\qquad
\epsilon_f:=b\epsilon^2,
\]
where $b>0$ is a constant independent of $d$
and $\epsilon$. Choose
\[
0<b\le
\min\left\{
1,\frac{C_h\Delta_p}{8u}
\right\}.
\]

Then
\[
\epsilon_g
=\epsilon_{g,1}+\epsilon_{g,2}
=\frac{1+\theta}{2}\eta_g\epsilon
<\eta_g\epsilon.
\]
Moreover, since $h(\epsilon)=C_h\epsilon^2$,
\[
\frac{4u\epsilon_f}{h(\epsilon)}
=\frac{4ub}{C_h}
\le\frac{\Delta_p}{2}.
\]

\paragraph{Probability of a true iteration.}
Set
\[
\delta_1:=\frac{1-p_0}{3},
\qquad
\sigma:=
\sqrt{\frac{\delta_1\epsilon_{g,2}^2}
{dL_{22}^2}}.
\]
Choose a single batch size satisfying
\[
m_{\mathrm{ZO}}
\ge
\max\left\{
1,\,
\frac{3\nu^2}{(1-p_0)\epsilon_{g,1}^2},\,
\frac{6L_q^{2/q}}{(1-p_0)\epsilon_f^2},\,
\frac{8d^2L_{22}^2L_q^{2/q}}
{\delta_1^2\epsilon_{g,2}^4}
\right\}.
\]
By Lemma~3, the response-gradient estimator
satisfies the required accuracy condition
with probability at least $1-\delta_1$.
This batch size also satisfies the
requirement of Lemma~2.

By Lemma~1, the probability of a true
iteration satisfies
\[
\begin{aligned}
p
&:=1-
\frac{\nu^2}
{m_{\mathrm{ZO}}\epsilon_{g,1}^2}
-\delta_1
-\frac{2L_q^{2/q}}
{m_{\mathrm{ZO}}\epsilon_f^2}\\
&\ge
1-\frac{1-p_0}{3}
-\frac{1-p_0}{3}
-\frac{1-p_0}{3}\\
&=p_0.
\end{aligned}
\]
Thus all admissibility conditions hold.

\paragraph{Resulting parameter orders.}
The finite-difference radius is
\[
\sigma
:=\sqrt{\frac{\delta_1\epsilon_{g,2}^2}
{dL_{22}^2}}
=\Theta\left(\frac{\epsilon}{\sqrt d}\right).
\]
In summary, the parameters can be chosen as
\[
\epsilon_{g,1}=\Theta(\epsilon),
\qquad
\epsilon_{g,2}=\Theta(\epsilon),
\qquad
\epsilon_f=\Theta(\epsilon^2),
\]
with
\[
m_{\mathrm{ZO}}
=
\Theta\left(
\max\left\{
\epsilon^{-2},
d^2\epsilon^{-4}
\right\}
\right)
=
\Theta(d^2\epsilon^{-4}).
\]
All hidden constants are independent of $d$
and $\epsilon$.

\subsection{Proof of Proposition~\ref{prop:sample_complexity}}
\addcontentsline{apc}{subsection}{\protect\numberline{\thesubsection}Proof of Proposition~\ref{prop:sample_complexity}}

\begin{proof}
The parameter choices in the preceding subsection ensure
$\epsilon_g<\eta_g\epsilon$ and $p>p_u(\epsilon)$, with
\[
m_{\mathrm{ZO}}
=\Theta\left(d^2\epsilon^{-4}\right).
\]
Each iteration uses one base batch,
$d$ perturbed-deployment batches, and one trial batch,
so
\[
N_{\mathrm{iter}}
=(d+2)m_{\mathrm{ZO}}
=O\left(d^3\epsilon^{-4}\right).
\]
\end{proof}

\begin{corollary}[End-to-end guarantee for PTRO]
\label{cor:end_to_end}
Suppose Assumptions~\ref{assp:PR_smooth}--\ref{assp:nabla_x_2_R_lip}
hold, and let $\epsilon\in(0,1)$ be a target
stationarity tolerance.
Choose the parameters
\[
\epsilon_{g,1}=\Theta(\epsilon),
\qquad
\epsilon_{g,2}=\Theta(\epsilon),
\qquad
\epsilon_f=\Theta(\epsilon^2),
\qquad
\sigma=\Theta\left(\frac{\epsilon}{\sqrt d}\right),
\]
and
\[
m_{\mathrm{ZO}}
=\Theta\left(d^2\epsilon^{-4}\right),
\]
with constants chosen as in Appendix D.5.
Then the estimator-accuracy conditions required by
Theorem~\ref{thm:convergence} are satisfied.

Consequently, PTRO reaches an $\epsilon$-stationary point,
\[
\min_{0\le k\le t-1}
\|\nabla PR(X_k)\|
\le\epsilon,
\]
with the high-probability guarantee stated in
Theorem~\ref{thm:convergence} after
$t=O(\epsilon^{-2})$ iterations.
Each iteration requires $d+2$ population queries and
\[
(d+2)m_{\mathrm{ZO}}
=O\left(d^3\epsilon^{-4}\right)
\]
samples.
\end{corollary}

\subsection{Convergence result under special cases}
\addcontentsline{apc}{subsection}{\protect\numberline{\thesubsection}Convergence Result under Special Cases}
\label{sec:converge_for_special_cases}
For the following special cases, we can obtain almost-sure finite-time convergence result.

\begin{corollary}[Uniformly bounded loss]
\label{crlr:uniform_bdd_loss}
Assume the conditions of Theorem~\ref{thm:convergence} hold and
$0\le \ell(z;x)\le M_1$ for all $z,x$. Then, for any $s>0$,
\[
\mathbb P\{T_\epsilon\le t\}
\ge
1-
\exp\!\left(
-\frac{(p-\widehat p)^2}
{2\max\{p,1-p\}^2}\,t
\right)
-
\exp\!\left(
-\frac{s^2t}{2M_1^2}
\right),
\]
for any
$\widehat p\in
\left(
p_u+\frac{us}{h(\epsilon)},\,p
\right)$
and any
$t>
\frac{R}
{\widehat p-p_u-us/h(\epsilon)},$
where
$R=
\frac{uZ_0}{h(\epsilon)}
+
\frac{u}{u+1}
\max\left\{
0,\,
\left\lceil
\frac{\ln(\alpha_0/\bar\alpha)}
{\ln(1/\gamma_{\mathrm{dec}})}
\right\rceil
\right\}.$

In particular, $T_\epsilon<\infty$ almost surely.
\end{corollary}

\begin{proof}
Define
\[
Y_k
:=
(E_k+E_k^+)
-
\mathbb E[E_k+E_k^+\mid\mathcal F_{k-1}],
\qquad
S_t:=\sum_{k=0}^{t-1}Y_k.
\]
Then
\[
\mathbb E[Y_k\mid\mathcal F_{k-1}]=0,
\]
so $(Y_k,\mathcal F_k)$ is a martingale-difference sequence.

Since $0\le \ell(z;x)\le M_1$, both empirical and population risks
belong to $[0,M_1]$, and hence
\[
0\le E_k\le M_1,
\qquad
0\le E_k^+\le M_1.
\]
Therefore,
\[
0\le E_k+E_k^+\le 2M_1.
\]
Conditionally on $\mathcal F_{k-1}$, $Y_k$ is thus centered and lies
in an interval of length at most $2M_1$. Hoeffding--Azuma therefore gives
\[
\mathbb P(S_t>st)
\le
\exp\!\left(
-\frac{2s^2t^2}{t(2M_1)^2}
\right)
=
\exp\!\left(
-\frac{s^2t}{2M_1^2}
\right).
\]

Moreover, by the conditional ZO mean bound,
\[
\mathbb E[E_k+E_k^+\mid\mathcal F_{k-1}]
\le 2\epsilon_f,
\]
and hence
\[
\sum_{k=0}^{t-1}
(2\epsilon_f+E_k+E_k^+)
\le
4\epsilon_f t+S_t.
\]
Thus,
\[
\mathbb P\!\left(
\sum_{k=0}^{t-1}
(2\epsilon_f+E_k+E_k^+)
>
(4\epsilon_f+s)t
\right)
\le
\exp\!\left(
-\frac{s^2t}{2M_1^2}
\right).
\]

Combining this bound with the true-iteration concentration argument in
Theorem~\ref{thm:convergence} gives
\[
\mathbb P\{T_\epsilon\le t\}
\ge
1-
\exp\!\left(
-\frac{(p-\widehat p)^2}
{2\max\{p,1-p\}^2}\,t
\right)
-
\exp\!\left(
-\frac{s^2t}{2M_1^2}
\right).
\]

By the Borel–Cantelli lemma, the probability of $T_\epsilon >t$ occurring infinitely often is $0$, $T_\epsilon<\infty$ almost surely.
\end{proof}

\begin{corollary}[Gaussian]
Assume that for every fixed $x\in\mathcal X$, the map
$\ell(\cdot;x)$ is $L_\ell$-Lipschitz in $z$, and that
\[
z\sim\mathcal N(A(x),\sigma^2I_d).
\]
Then, for any $s>0$, there exists an absolute constant $c>0$ such that
\[
\mathbb P\{T_\epsilon\le t\}
\ge
1-
\exp\!\left(
-\frac{(p-\widehat p)^2}
{2\max\{p,1-p\}^2}\,t
\right)
-
\exp\!\left(
-c\,t
\min\left\{
\frac{ms^2}{L_\ell^2\sigma^2},
\frac{\sqrt m\,s}{L_\ell\sigma}
\right\}
\right),
\]
for any
$\widehat p\in
\left(
p_u+\frac{us}{h(\epsilon)},\,p
\right),$
and any
$t>
\frac{R}{\widehat p-p_u-us/h(\epsilon)},$

where 
$R=
\frac{uZ_0}{h(\epsilon)}
+
\frac{u}{u+1}
\max\left\{
0,\,
\left\lceil
\frac{\ln(\alpha_0/\bar\alpha)}
{\ln(1/\gamma_{\mathrm{dec}})}
\right\rceil
\right\}.$
In particular, $T_\epsilon<\infty$ almost surely.
\label{crlr:Gaussian}
\end{corollary}

\begin{proof}
Consider first a generic function-value query at a point
$(x_1,x_2)$ that is fixed conditional on the history available before
its minibatch is drawn. Write
\[
Z=A(x_2)+\sigma U,
\qquad
U\sim\mathcal N(0,I_d),
\]
and define
\[
g(u):=\ell(A(x_2)+\sigma u;x_1).
\]
Since $\ell(\cdot;x_1)$ is $L_\ell$-Lipschitz in $z$, the map $g$ is
$\sigma L_\ell$-Lipschitz in $u$. Hence Gaussian concentration gives
\[
\mathbb P\!\left(
|g(U)-\mathbb Eg(U)|\ge r
\right)
\le
2\exp\!\left(
-\frac{r^2}{2L_\ell^2\sigma^2}
\right),
\qquad r\ge0.
\]
Therefore, for an empirical average based on $m$ independent samples,
the signed estimation error
\[
\Delta(x_1,x_2)
:=
f(x_1,x_2)-R(x_1,x_2)
\]
satisfies
\[
\|\Delta(x_1,x_2)\|_{\psi_2}
\le
C\frac{L_\ell\sigma}{\sqrt m}
\]
for an absolute constant $C>0$.

Since $x\mapsto |x|$ is $1$-Lipschitz, the corresponding centered
absolute error also satisfies
\[
\left\|
E-\mathbb E[E\mid\mathcal H]
\right\|_{\psi_2\mid\mathcal H}
\le
C\frac{L_\ell\sigma}{\sqrt m},
\]
where $\mathcal H$ denotes the history available immediately before
that function-value minibatch is drawn.

For the current-point evaluation, take
$\mathcal H=\mathcal F_{k-1}$ and define
\[
U_k
:=
E_k-\mathbb E[E_k\mid\mathcal F_{k-1}].
\]
For the trial-point evaluation, let $\mathcal G_k$ denote the
within-iteration history after $X_k^+$ has been constructed but before
the trial minibatch is drawn, and define
\[
V_k
:=
E_k^+-\mathbb E[E_k^+\mid\mathcal G_k].
\]
Then
\[
\mathbb E[U_k\mid\mathcal F_{k-1}]=0,
\qquad
\mathbb E[V_k\mid\mathcal G_k]=0,
\]
and both increments satisfy the uniform conditional sub-Gaussian bound
\[
\|U_k\|_{\psi_2\mid\mathcal F_{k-1}},
\;
\|V_k\|_{\psi_2\mid\mathcal G_k}
\le
C\frac{L_\ell\sigma}{\sqrt m}.
\]

Ordering the current-point and trial-point evaluations according to
their sampling times therefore gives a martingale-difference sequence
with conditionally sub-exponential increments satisfying
\[
\|\cdot\|_{\psi_1}
\le
C\frac{L_\ell\sigma}{\sqrt m}.
\]
Consequently, a Bernstein inequality for martingale differences gives
\[
\mathbb P\!\left(
\sum_{k=0}^{t-1}(U_k+V_k)>st
\right)
\le
\exp\!\left(
-c\,t
\min\left\{
\frac{ms^2}{L_\ell^2\sigma^2},
\frac{\sqrt m\,s}{L_\ell\sigma}
\right\}
\right).
\]

By the individual conditional ZO mean bounds,
\[
\mathbb E[E_k\mid\mathcal F_{k-1}]
\le\epsilon_f,
\qquad
\mathbb E[E_k^+\mid\mathcal G_k]
\le\epsilon_f.
\]
Hence, pathwise,
\[
E_k+E_k^+-2\epsilon_f
\le
U_k+V_k.
\]
Therefore,
\[
\left\{
\sum_{k=0}^{t-1}
(E_k+E_k^+-2\epsilon_f)>st
\right\}
\subseteq
\left\{
\sum_{k=0}^{t-1}(U_k+V_k)>st
\right\},
\]
and thus
\[
\mathbb P\!\left(
\sum_{k=0}^{t-1}
(E_k+E_k^+-2\epsilon_f)>st
\right)
\le
\exp\!\left(
-c\,t
\min\left\{
\frac{ms^2}{L_\ell^2\sigma^2},
\frac{\sqrt m\,s}{L_\ell\sigma}
\right\}
\right).
\]

Combining this accumulated-error bound with the true-iteration
concentration argument in Theorem~\ref{thm:convergence} gives
\[
\mathbb P\{T_\epsilon\le t\}
\ge
1-
\exp\!\left(
-\frac{(p-\widehat p)^2}
{2\max\{p,1-p\}^2}\,t
\right)
-
\exp\!\left(
-c\,t
\min\left\{
\frac{ms^2}{L_\ell^2\sigma^2},
\frac{\sqrt m\,s}{L_\ell\sigma}
\right\}
\right).
\]

For fixed $s>0$ and admissible $\widehat p$, both failure terms tend to
zero as $t\to\infty$. Therefore,
\[
\mathbb P(T_\epsilon=\infty)
=
\lim_{t\to\infty}\mathbb P(T_\epsilon>t)
=0.
\]
Hence $T_\epsilon<\infty$ almost surely.
\end{proof}

\section{Additional Details and Experiments}
\addcontentsline{apc}{section}{\protect\numberline{\thesection}Additional Details and Experiments}

\label{appdx_sec:experiment_detials}
All experiments were conducted on a MacBook Pro with an Apple M1 Pro chip (CPU) and 16 GB memory, using standard Python libraries. No GPU or specialized 
hardware was required. Each individual experimental run completes within a few minutes. The total 
compute required for all reported experiments is on the order of several hours. The overall research project did not require substantially more computing beyond the experiments reported in the paper.

Details are as follows.
\subsection{Strategic classification with Heterogeneous Responses}
\label{appdix:ablation_setup}
\addcontentsline{apc}{subsection}{\protect\numberline{\thesubsection}Strategic classification with Heterogeneous Responses}

In this experient, we omit Plug-in because its performance depends critically on the
chosen response-model atlas, i.e., the set of candidate response maps
$D(\cdot)$. If the atlas contains the true response model, the problem
largely reduces to response estimation followed by optimization,
whereas a misspecified atlas introduces an additional modeling error
that is absent from our setting. For DFO$(\lambda)$, since the population's response exhibits no Markov structure in our setting, use a plain two-point estimate with 
$O(1)$ samples per iteration.

\paragraph{Population response model.}
\begin{itemize}
    \item \textbf{Linear.} All individuals follow the same linear response rule. With an unknown-to-the-learner
    sensitivity matrix $A$,  utility $u(x,z) = -\langle Ax, z \rangle$
    and quadratic manipulation cost
    $\frac{1}{2\rho}\|z - z_0\|_2^2$, the best response is
    $$
        z^\star(x)
        =
        \arg\max_z
        \left\{
            -\langle Ax, z \rangle
            -\frac{1}{2\rho}\|z - z_0\|_2^2
        \right\}
        =
        z_0 - \rho A x,
    $$
    
    where $z_0$ denotes the baseline features and $\rho>0$
    controls response strength. This expression applies to
    the manipulable coordinates.

   \item \textbf{Mixture-linear (MixL).}
    The population is divided into two groups according to the
    true label $y_i\in\{0,1\}$. The groups share the same response
    map $A$ but respond in opposite directions:
    \[
        z_i^\star(x)
        =
        z_{0,i}+\rho\,s(y_i)Ax,
        \qquad s(y_i):=2y_i-1.
    \]
    Thus, individuals with $y_i=1$ shift by $+\rho Ax$, 
    those with $y_i=0$ shift by $-\rho Ax$. This setting introduces
    label-dependent heterogeneity.
\item \textbf{Mixture-nonlinear (MixN).}
    We keep the two  groups and replace the
    linear response with a saturating nonlinear rule:
    \[
        z_i^\star(x)
        =
        z_{0,i}
        +\rho\,s(y_i)\,C
        \tanh\!\left(\frac{Bx}{\tau}\right).
    \]
    Here, $\tanh$ acts coordinatewise,
    $B$ is a sensitivity matrix, and $m$ is the number of manipulable
    features. The parameter $\tau>0$ controls the input scale,
    while $C>0$ controls the saturation amplitude.
\end{itemize}
All three add independent Gaussian noise to the shift. $A$, $B$, $\rho$, $\sigma$, and $C$ are fixed but never revealed to any method; every algorithm observes only samples from the resulting population.

\paragraph{Sample budget matching.} We compare the methods under the same total sample budget.

We fix PTRO's $T$-iteration sample cost as a shared budget and choose the number of iterations for
each baseline so that its total query cost is approximately the same. For per iteration sample complexity,  PTRO spends $(d+2)m_{\mathrm{zo}}$  queries per iteration, DSA-PGM spends $b(1+2d)$, PerfGD  spends $m_{\mathrm{zo}}(d+1)$, DFO($\lambda$) spend $m_{\mathrm{zo}}$

For Two-Stage, which is not iterative updates, we choose the
number of exploration deployments matching the total sample budget.

\paragraph{Dataset.} We use the \emph{Give Me Some Credit} credit-scoring dataset~\citep{givemesomecredit2012} and the prediction task from~\citep{ding2021retiring},  built from the 2018 1-Year ACS PUMS for California~\citep{acs_pums} . The datasets provide the initial population. 

\paragraph{Hyperparameters.} 
\begin{itemize}
    \item \textbf{PTRO}: initial trust-region radius $\alpha_0=0.1$,
    finite-difference scale $\sigma$ selected from $\sigma\in\{0.1,0.5,0.8,1.2\}$, acceptance thresholds
    $\eta_1=0.2$ and $\eta_2=0.3$, and radius update factors
    $\gamma_{\mathrm{inc}}=2.0$ and $\gamma_{\mathrm{dec}}=0.5$.

    \item \textbf{Two-Stage}: exploration perturbation scale $0.2$.
    The optimization phase uses Adam, with the learning rate selected best from
    $\{0.01,0.03,0.05\}$.

    \item \textbf{DSA-PGM}: finite-difference scale $\delta=0.2$ for
    estimating the response Jacobian, with the learning rate selected best from
    $\{0.01,0.03,0.05\}$.

    \item \textbf{PerfGD}: trailing-window length $H=10$ and learning rate selected from
    $\{0.01,0.03,0.05\}$.

    \item \textbf{DFO($\lambda)$}: finite-difference scale $\sigma$ selected from
    $\delta\in\{0.1,0.2,0.3,0.5,0.8,1.2\}$ for the
    full-gradient estimate, with the learning rate selected best from
    $\{0.001,0.003,0.01,0.03\}$.

    \item \textbf{GZO-NS} and \textbf{GZO-HS}: smoothing
    parameter $\mu_k=\max(0.2\cdot 0.999^k,\,0.02)$, guide-mixing weight
    $\alpha_{k+1}=1-0.999\,(1-\alpha_k)$ with $\alpha_0=0.1$, and
    $\Sigma_0=I/d$. GZO-NS draws a fresh batch of $n=m$ samples per iteration for the
    guide gradient; GZO-HS instead reuses the samples of the last $p+1=2$
    iterations. The step size $\beta$ is constant and selected from
    $\{0.03,0.1,0.3,1.0\}$.
\end{itemize}

For each hyperparameter setting, our results are averaged over $20$ independent trials, and we report the mean $\pm$ standard error in the figures.

\subsection{Robustness to Population Heterogeneity}
\label{appdx_subsec:Robustness to Population Heterogeneity}
\addcontentsline{apc}{subsection}{\protect\numberline{\thesubsection}Robustness to Population Heterogeneity}
\paragraph{Setup.}
The previous mixture experiments show that heterogeneous population responses can challenge methods that rely on a prespecified response model. 
To study this effect more systematically, we consider a controlled two-group mixture and sweep the mixture fraction $\beta\in[0,1]$.

When $x$ is deployed, individual $i$
moves its mutable block $z_i^{\mathrm{m}}$ to
$z_i^{\mathrm{m}} + \rho\,\delta_i(x) + \varepsilon_i$, with $\rho=5$ and
$\varepsilon_i\sim\mathcal{N}(0,0.05^2 I)$. Each individual has a fixed type
$t_i\in\{0,1\}$:
\[
\delta_i(x)=
\begin{cases}
A x, & t_i=0,\\
A x_0 + A'(x-x_0), & t_i=1,
\end{cases}
\]
where $A,A'$ have i.i.d.\ $\mathcal{N}(0,1/9)$
entries, are drawn independently, and are held fixed. When $\beta=0$ and $0$, the population is homogeneous and is well aligned with a single linear response model. Thus, model misspecification is smallest at the two endpoints and is induced by the coexistence of heterogeneous response types in the intermediate mixture regimes.

We compare PTRO with Two-Stage, DSA-PGM, PerfGD, and DFO($\lambda$). We report the final stationarity measure $\|\nabla PR(x_T)\|$ and the number of samples required to reach the target
$\|\nabla PR(x)\| \le 0.02.$

Runs that do not reach the target within the sampling budget are capped at the budget.

\begin{figure}[t]
    \centering
\includegraphics[width=\linewidth]{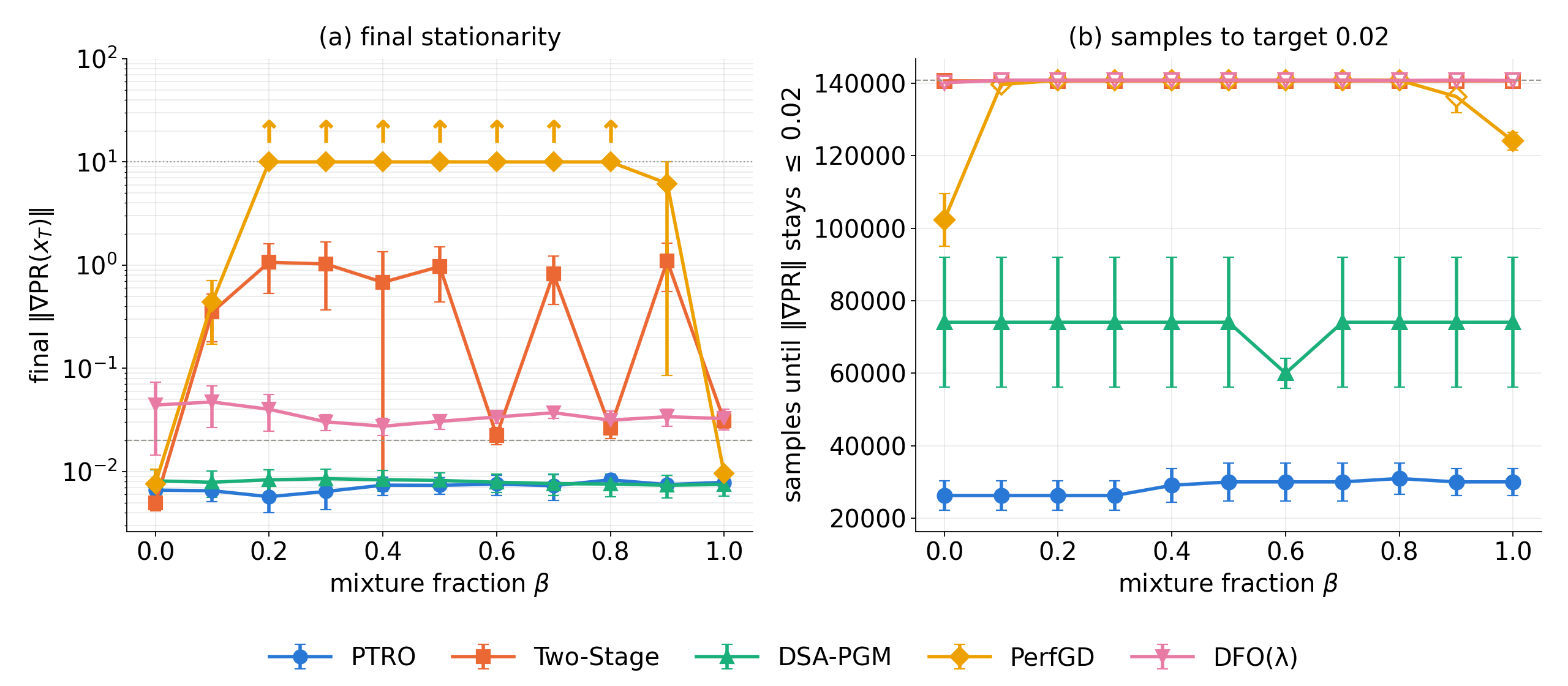}

    \caption{
    Robustness to population heterogeneity.}
\label{fig:robut_to_heterogeneity}

\end{figure}

\paragraph{Results.}
As shown in Figure~\ref{fig:robut_to_heterogeneity}, PTRO remains stable across the full range of mixture fractions. Its final stationarity stays around $10^{-2}$, and the number of samples required to reach the target changes only mildly as $\beta$ increases.

In contrast, the response-model-based methods are more sensitive to population heterogeneity. Two-Stage and PerfGD perform well near the homogeneous setting, but become unstable or fail to reach low stationarity for many intermediate mixture fractions. DSA-PGM is more stable, but requires more samples than PTRO to reach the same stationarity target. DFO($\lambda$) is relatively insensitive to $\beta$, but converges to a higher stationarity level and generally does not reach the target $0.02$.

\subsection{Effect of  Response Distribution}
\label{appdx_subsec:effect_of_response_distribution}
\addcontentsline{apc}{subsection}{\protect\numberline{\thesubsection}Effect of  Response Distribution}
\paragraph{Problem}

Using the same dataset as in the previous experiment, we examine how
PTRO behaves as the population response becomes increasingly
heterogeneous. We examine two dimensions: its
magnitude and the number of distinct response types.

For magnitude, we define an individual-specific response matrix
  $A(x_i) = A + \omega\, h_i\, B,$
where  $h_i$ is a centered nonlinear score that varies across individuals, $\omega$ controls the magnitude of the heterogeneous response
component, larger $\omega$ produces larger deviations from the baseline response matrix $A$.
Also, we vary $K$ to control how many distinct response behaviors are present
in the population, ranging from a single shared response ($K=1$) to
fully individualized responses ($K=n$).

\paragraph{Setup.}
We use the Give Me Some Credit environment with the same logistic loss. Under deployment of $x$, individual $i$ changes the
manipulatable features according to
\[
    \Delta_i(x)=\rho A_i x+\xi_i,
    \qquad
    \rho=0.5,
    \qquad
    \xi_i\sim\mathcal N(0,0.05^2 I_M).
\]
We consider four cases of group dividing:
\begin{align*}
    K=1: A_i=\bar A; \quad
    K=2:\ A_i=\bar A_{g_i},\ g_i\in\{1,2\}; 
    \\
    K=5:\ A_i=\bar A_{g_i},\ g_i\in\{1,\cdots,5\}; \quad
    K=n:\ A_i=A(z_i).
\end{align*}

All four settings are derived from the same individual-level map
\[
    A(z_i)=A+\omega h_i B,
\]
where $A,B\in\mathbb R^{M\times d}$ have i.i.d.\
$\mathcal N(0,1/d)$ entries, and
\[
    h_i=
    \tanh\!\left(
        \frac{\phi(x_i)^\top\theta_0-\mu}{s}
    \right)-c,
    \qquad
    \mathbb E[h]=0.
\]
For $K=2$, individuals are split at the median of $h$ and
$\bar A_g=\mathbb E[A(z)\mid g]$; for $K=1$,
$\bar A=\mathbb E[A(z)]=A$. Hence all three settings have the same
population-average response, while $K=1,2,n$ retain increasing amounts of
individual heterogeneity. We use $\kappa\in\{1,3\}$ to vary its strength.

\begin{figure}[t]
    \centering

    \begin{subfigure}{0.32\textwidth}
        \centering
        \includegraphics[height=3cm,width=\linewidth]{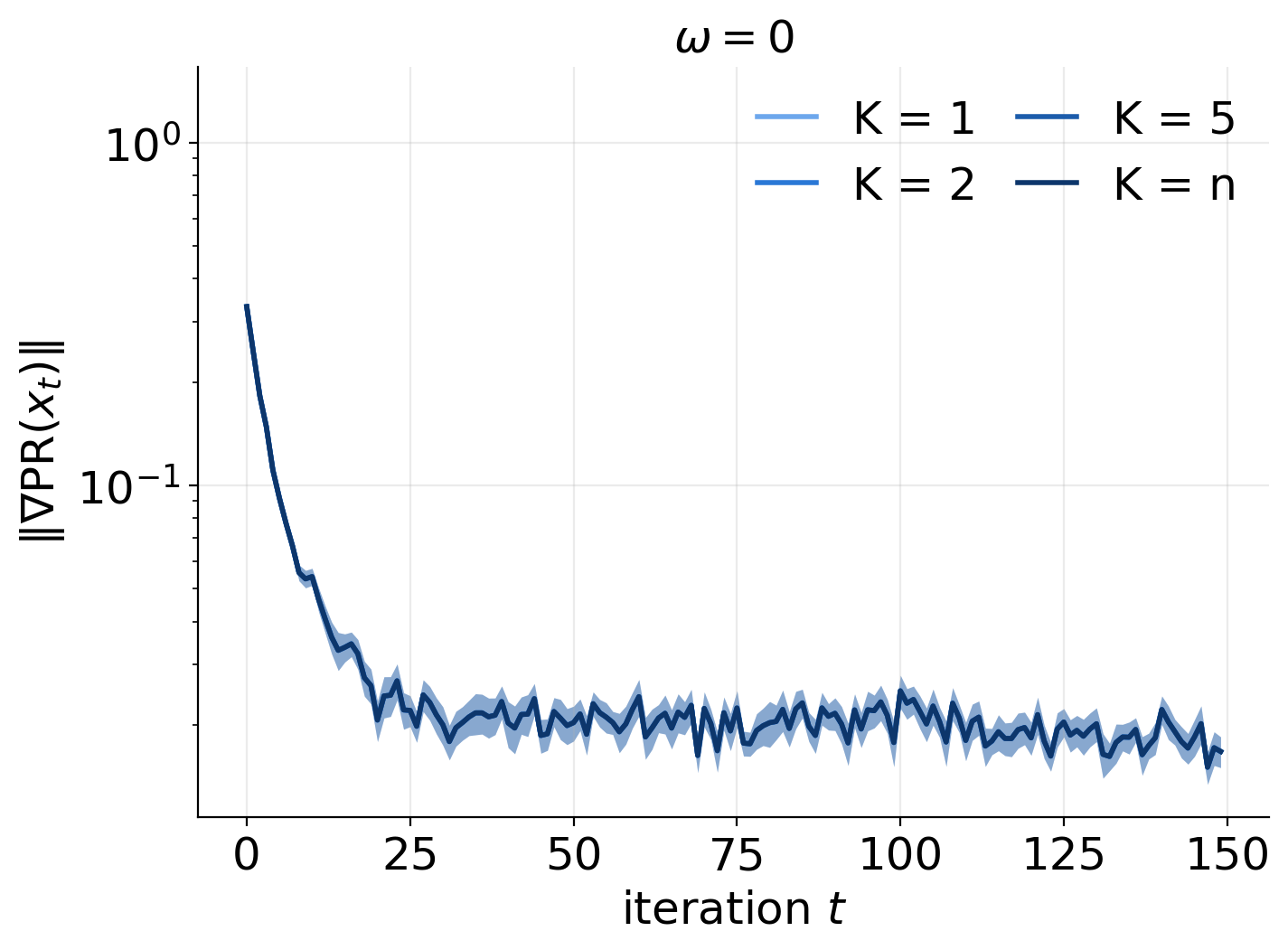}

    \end{subfigure}
    \hfill
    \begin{subfigure}{0.32\textwidth}
        \centering
        \includegraphics[height=3cm,width=\linewidth]{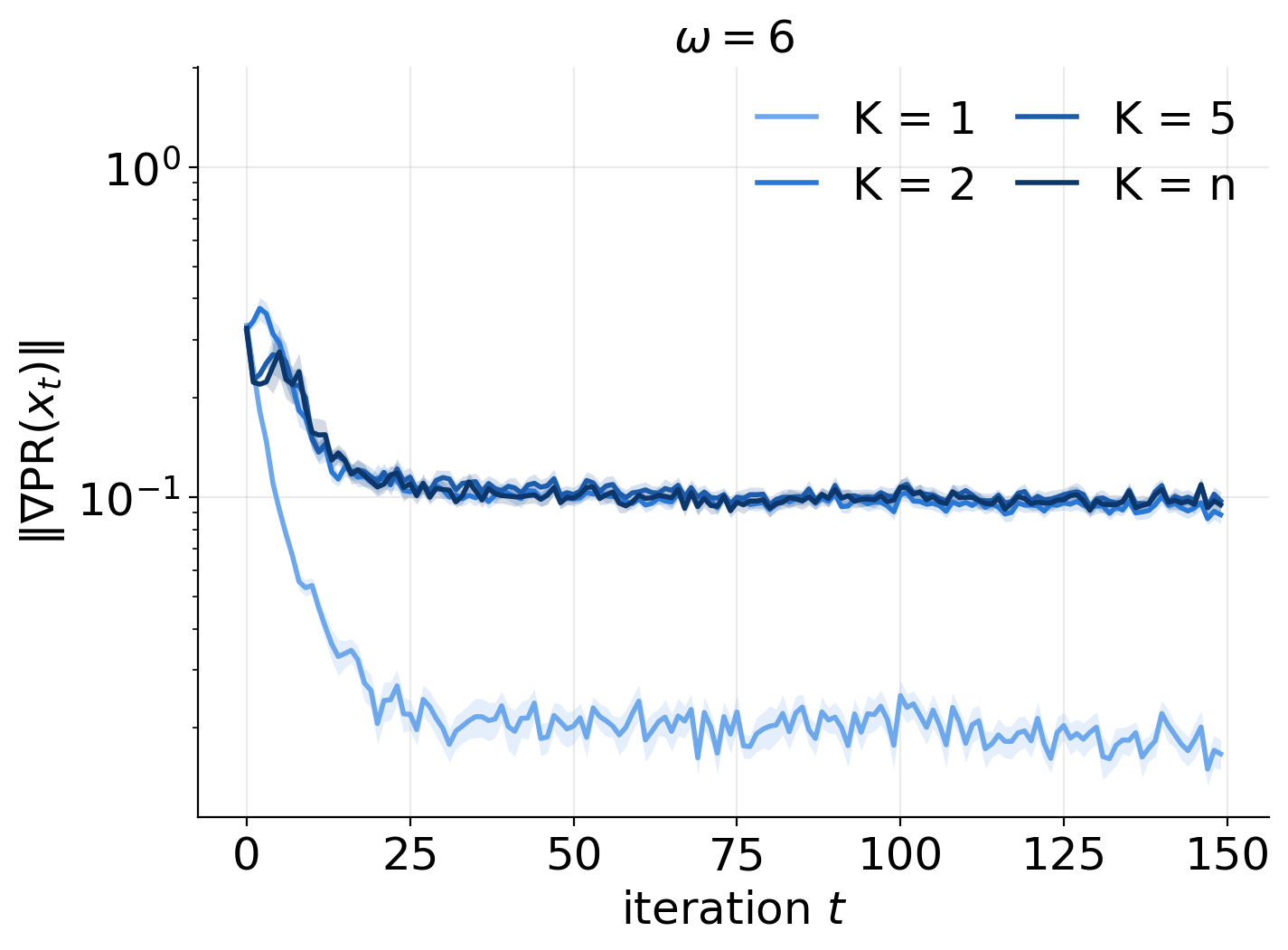}
    \end{subfigure}
    \hfill
    \begin{subfigure}{0.32\textwidth}
        \centering
        \includegraphics[height=3cm,width=\linewidth]{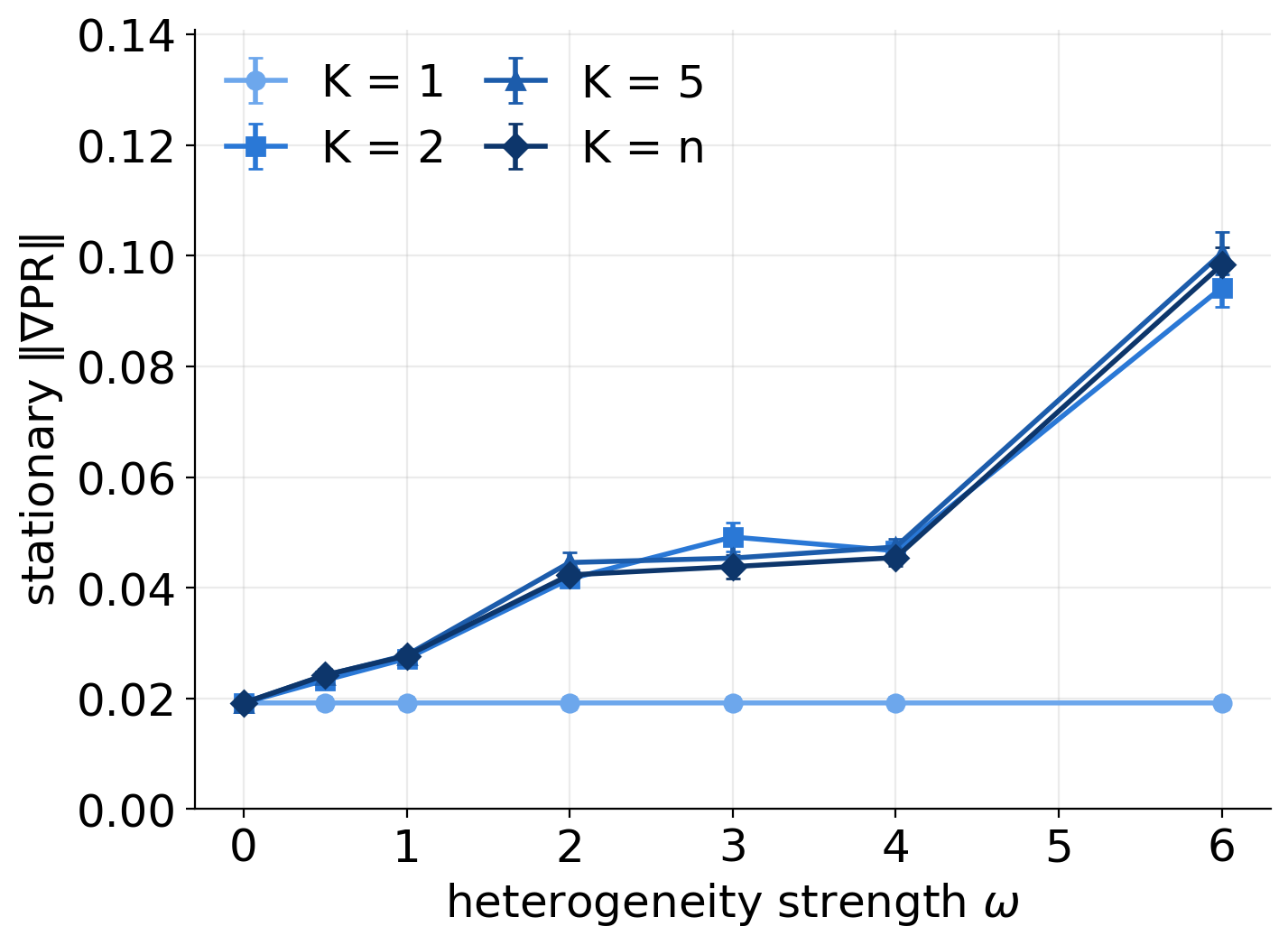}
    \end{subfigure}

    \caption{Effect of response heterogeneity on PTRO.}
    \label{fig:effect_of_response_dist}
\end{figure}

\paragraph{Result.}
Figure~\ref{fig:effect_of_response_dist} shows, as $\omega$ increases,  stronger response
heterogeneity makes optimization more difficult. For $K=1$, the heterogeneous component is largely cancels because $h_i$ is centered. For $K>1$, this cancellation no
longer holds within each group. The curves for $K>1$ remain close, suggesting that the magnitude of heterogeneity has a larger effect than the number of distinct response types. This behavior is
consistent with Proposition~\ref{prop:lower_bd_for_eps}: stronger
heterogeneity can reduce the accuracy of the response-gradient estimate, so maintaining the same stationarity level may require larger
minibatches.

\subsection{Ablation Study}
\label{appdix:ablation_setup}
\addcontentsline{apc}{subsection}{\protect\numberline{\thesubsection}Ablation Study}

We study the contributions of using known loss function and the adaptive
trust-region mechanism separately.

\paragraph{Exploiting the known loss.}
We compare PTRO with a black-box trust-region method (TR-fullGrad) that
approximates the entire performative risk using a
finite-difference gradient estimator. 
Both variants use 
the same quadratic correction and trust-region update rules.
This comparison evaluates the benefit
of exploiting the known loss rather than treating the
performative risk as a black-box objective.

\begin{figure}[t]
    \centering

    \begin{subfigure}[t]{0.32\textwidth}
        \centering
        \includegraphics[width=\linewidth]{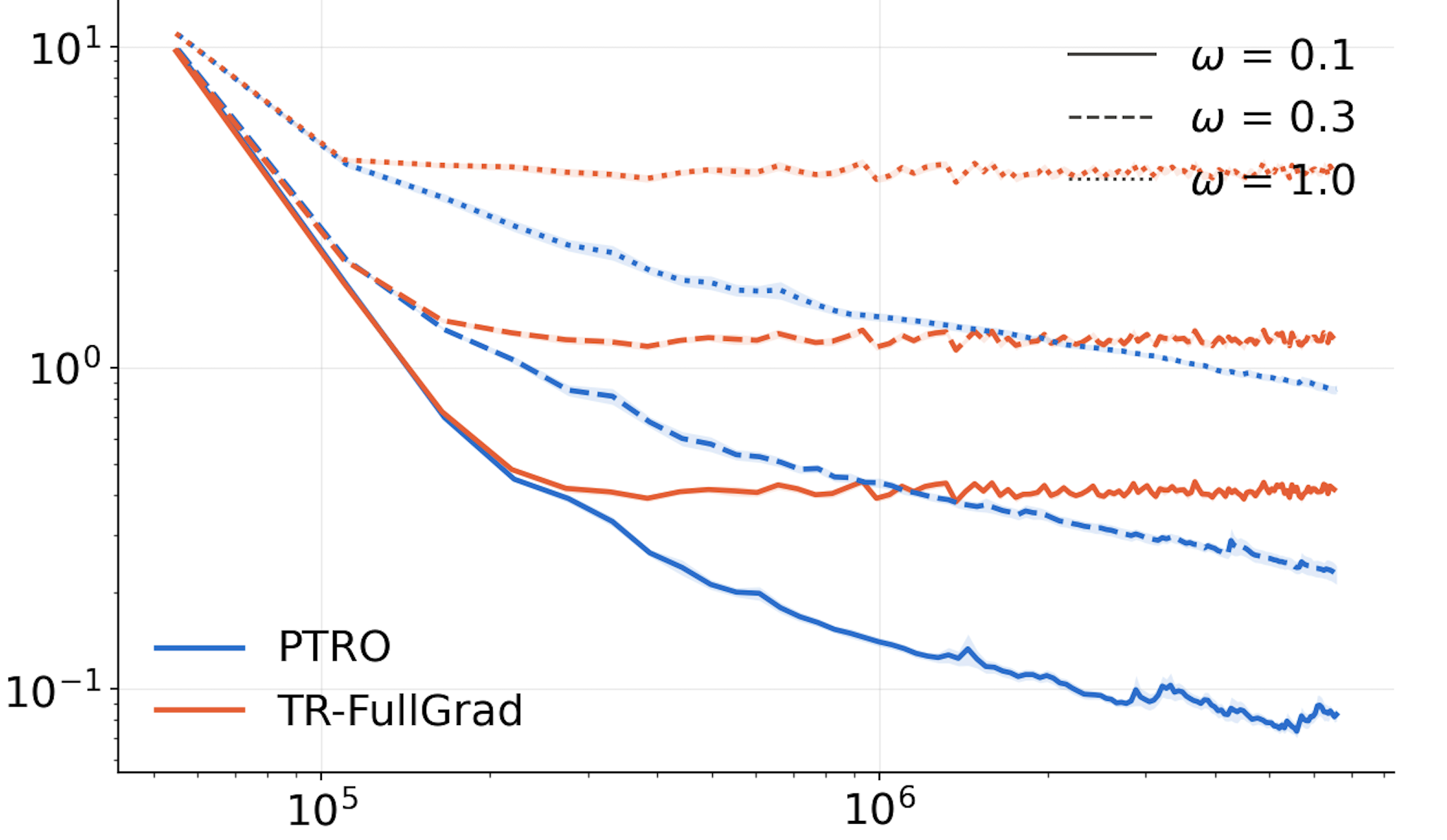}
        \caption{Exploiting the known loss}
        \label{fig:ablation_known_loss}
    \end{subfigure}
    \hfill
    \begin{subfigure}[t]{0.32\textwidth}
        \centering
        \includegraphics[width=\linewidth]{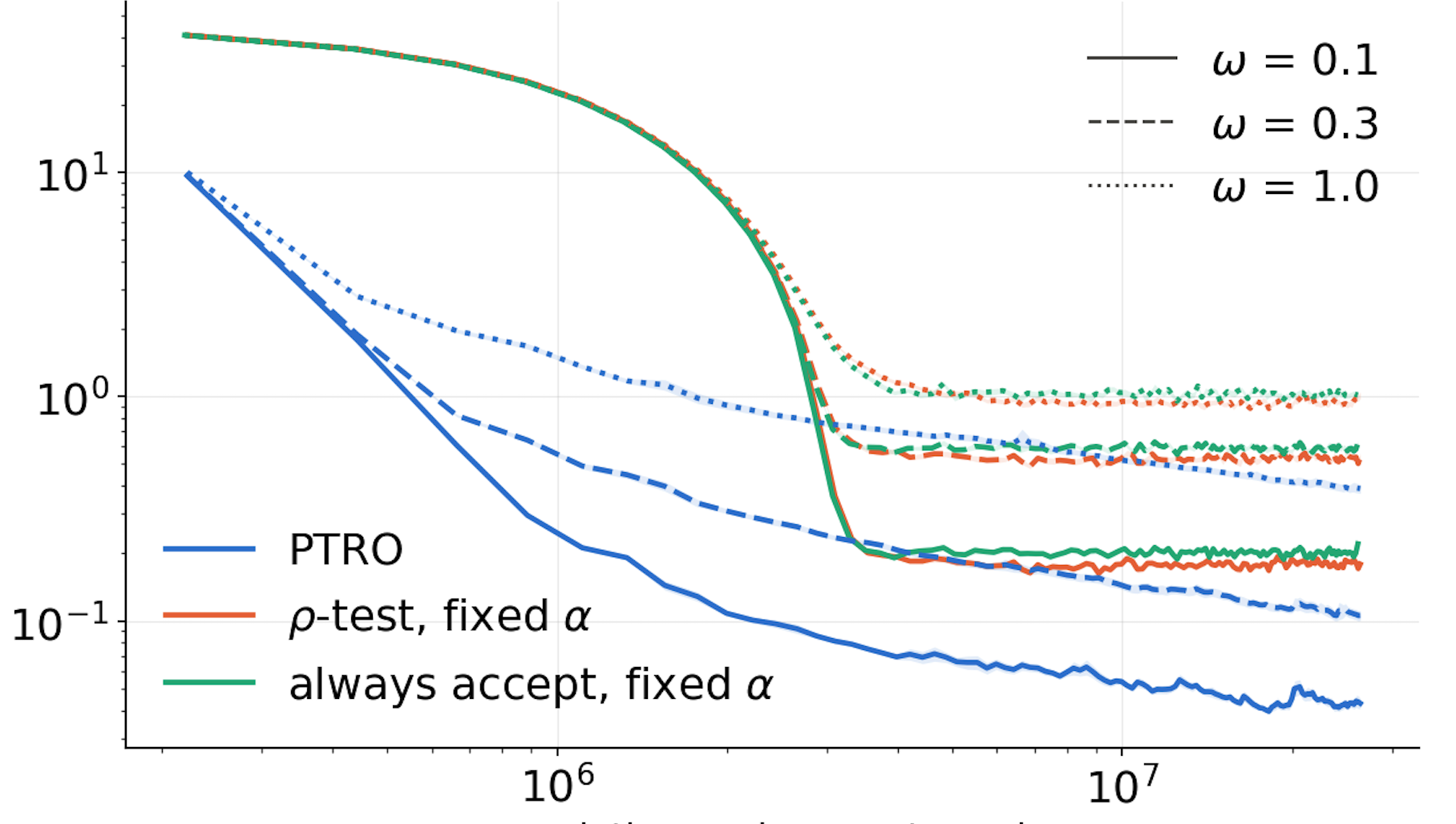}
        \caption{Trust-region mechanism ($\alpha=0.1$)}
        \label{fig:ablation_alpha_small}
    \end{subfigure}
    \hfill
    \begin{subfigure}[t]{0.32\textwidth}
        \centering
        \includegraphics[width=\linewidth]{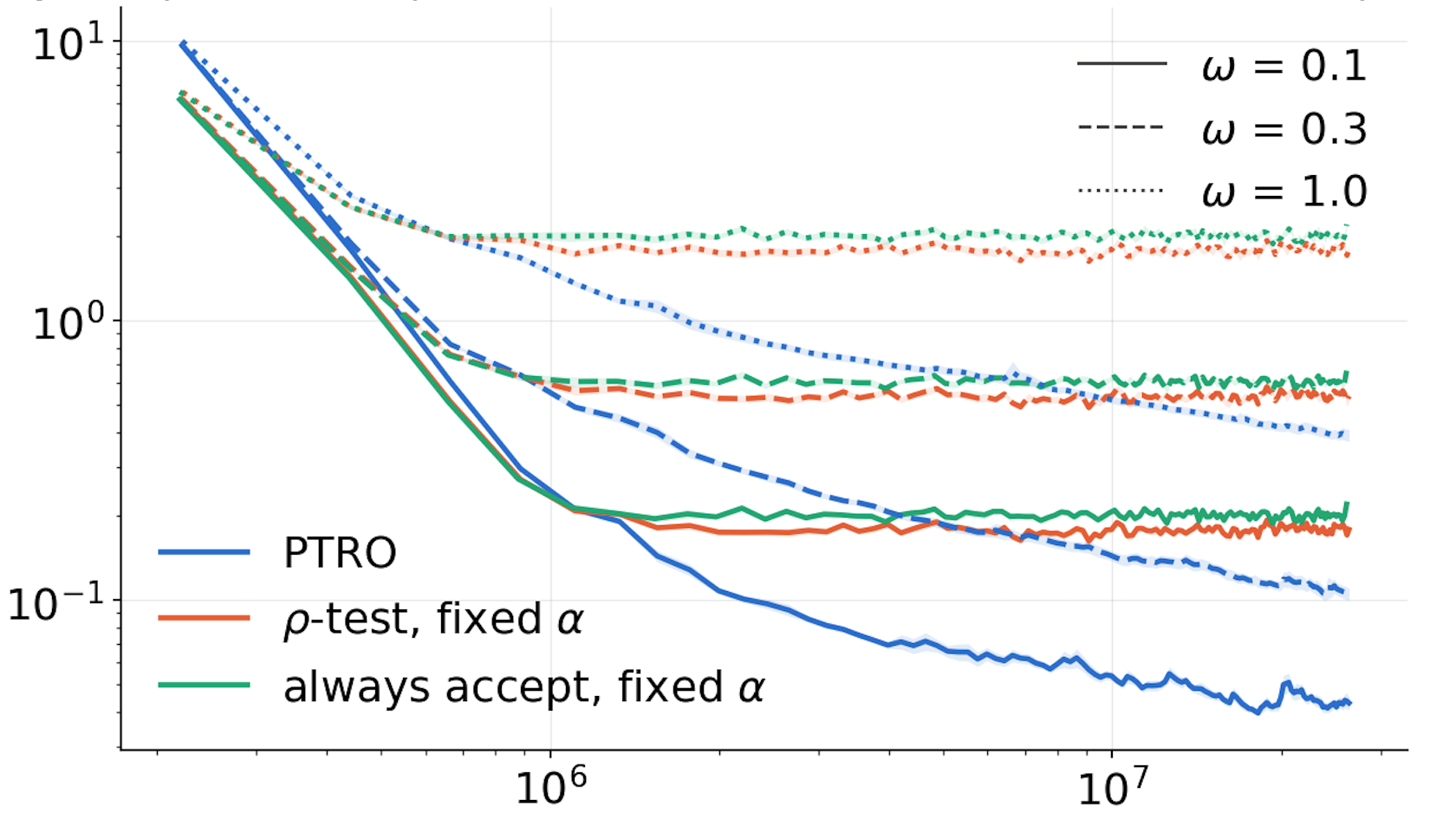}
        \caption{Trust-region mechanism ($\alpha=1.0$)}
        \label{fig:ablation_alpha_large}
    \end{subfigure}

    \caption{
    Ablation study of the main components of PTRO.
    The left panel compares PTRO with a black-box trust-region method
    that estimates the full performative-risk gradient using finite
    differences. The middle and right panels isolate the effects of the
    acceptance test and trust-region radius adaptation under two fixed
    radii. The $x$-axis shows cumulative environment queries and the
    $y$-axis shows $\|\nabla PR(x_k)\|$.
    }
    \label{fig:ablation_study}
\end{figure}

\paragraph{Acceptance and trust-region adaptation.}
We compare PTRO with two variants: one keeps the acceptance
test but fixes the trust-region radius ($\rho$-test, fixed $\alpha$ in the Figure~\ref{fig:ablation_study}), and the other uses
the same fixed radius and always accepts the proposed step(always accept, fixed $\alpha$ in the Figure~\ref{fig:ablation_study}).
All three methods use the same local-approximation construction and
sampling configuration. Comparing PTRO with the first variant
examines the effect of radius adaptation; comparing the two
fixed-radius variants isolates the effect of the acceptance test. We test with $\alpha =0.1, 1.0$.

\paragraph{Setup.}
With the weighted squared loss and a linear-Gaussian response,
\begin{equation}
\ell(z;x)=\tfrac12 (x-z)^\top Q (x-z),
\qquad
z = b + Ax + \xi,
\qquad
\xi\sim\mathcal N(0,\omega^2 I),
\label{eq:ablation_problem}
\end{equation}
writing $u(x):=(I-A)x-b$ gives
\begin{equation}
PR(x)=\tfrac12 u^\top Q u+\tfrac{\omega^2}{2}\operatorname{tr}(Q),
\qquad
\nabla PR(x)=(I-A)^\top Q u,
\qquad
x^\star=(I-A)^{-1}b,
\label{eq:ablation_closed_form}
\end{equation}
The algorithms observe only samples
, \eqref{eq:ablation_closed_form} is used for
evaluation only.

\paragraph{Configuration.}
$A$, $b$ and $Q$ are drawn once from a fixed seed and held constant across
trials.

\begin{center}
\begin{tabular}{ll}
\toprule
Dimension & $d=100$ \\
Response matrix & $A=0.2\,M/\|M\|_2$, $M\sim\mathcal N(0,1)^{d\times d}$ \\
Loss curvature & $Q=\operatorname{diag}(\text{logspace}(0,1,d))$, $\kappa(Q)=10$ \\
Baseline mean, initial decision & $b=\mathbf 1$, $x_0=0$ \\
Noise scale & $\omega\in\{0.1,0.3,1.0\}$ \\
Batch sizes & $m_{zo}\in\{16,64,256\}$ \\
FD step & $\sigma=0.003$ \\
Smoothness constants & $L_\ell=\|Q\|_2=10$, $\kappa_H=1$ \\
Trust region & $\alpha_0=0.2$, $\eta_1=0.1$, $\eta_2=0.5$,
  $\gamma_{\mathrm{inc}}=2.0$, $\gamma_{\mathrm{dec}}=0.5$ \\
Iterations, repetitions & $T=200$, 20 paired seeds \\
\bottomrule
\end{tabular}
\end{center}

\paragraph{Result.}
Figure~\ref{fig:ablation_known_loss} shows that exploiting the known loss can improve efficiency and the attainable stationarity level. The difference comes from how the two methods estimate the
performative-risk gradient. TR-FullGrad applies finite differences to the
entire objective, so estimation error from function evaluations affects both
the direct loss effect and the population-response effect. In contrast, PTRO
evaluates the direct effect through the gradient of the known loss and uses finite differences only for the unknown population-response component. This
removes an unnecessary source of zeroth-order estimation error.


Figure~\ref{fig:ablation_alpha_small} and~\ref{fig:ablation_alpha_large} highlight the importance of radius adaptation.
A small fixed radius slows early progress, while a large fixed radius leads to
a higher stationarity floor. Moreover, the acceptance test alone provides
limited improvement over always accepting, since rejection does not correct
the neighborhood size. PTRO instead couples acceptance with radius adaptation,
allowing larger steps when the local model is reliable and contracting the
radius when model agreement deteriorates. This yields consistently lower
stationarity error than the fixed-radius variants.

\subsection{Effect of Dimension and Sampling Budget}
\label{appdx_subsec:effects_of_dim_budget}
\addcontentsline{apc}{subsection}{\protect\numberline{\thesubsection}Effect of Dimension and Sampling Budget}
We vary the decision dimension $d$ and the total sampling budget per
PTRO iteration. 
 
\paragraph{Setup.}
We use the same quadratic performative-risk problem as in the preceding appendix experiments.

We consider dimensions $d\in\{10,50,100,200\}$ and three scaling regimes for the per-iteration sampling budget $N_{\mathrm{iter}}$: constant, $N_{\mathrm{iter}}\in\{2020,\,10100,\,20200\}$; linear, $N_{\mathrm{iter}}\in\{10d,\,50d,\,100d\}$; and quadratic, $N_{\mathrm{iter}}\in\{10d^2,\,50d^2,\,100d^2\}$. Each PTRO iteration performs $d+2$ risk evaluations (one at the current point, $d$ coordinate probes, and one at the trial point), and the budget is split evenly among them, with $m=\mathrm{round}\big(N_{\mathrm{iter}}/(d+2)\big)$ samples per evaluation. For each configuration, we run $T=30$ iterations over $10$ independent trials (seeds).

\paragraph{Result.}
The result is shown in Figure~\ref{appdx_subsec:effects_of_dim_budget}. First, increasing the per-iteration sampling budget improves the final
stationarity level. For every tested dimension, using more samples
reduces the final value of $\|\nabla PR\|$. This is consistent with
Proposition~\ref{prop:lower_bd_for_eps}: larger minibatches improve the
accuracy of the risk and response-gradient estimators, which permits a
smaller attainable stationarity tolerance.

Second, the number of iterations required for optimization progress is
comparatively stable across the tested dimensions. This is consistent
with Theorem~\ref{thm:convergence}, whose iteration-complexity bound has
no explicit polynomial dependence on $d$; the dimension dependence
enters primarily through the sampling requirements needed to achieve
the prescribed estimator accuracies.

Overall, the observed dimension dependence may be milder than the worst-case
sufficient sample complexity in Proposition~\ref{prop:sample_complexity},
suggesting quadratic per-iteration sampling as a practical choice..

\begin{figure}[t]
    \centering

    \begin{subfigure}{0.95\linewidth}
        \centering
        \includegraphics[width=\linewidth]
        {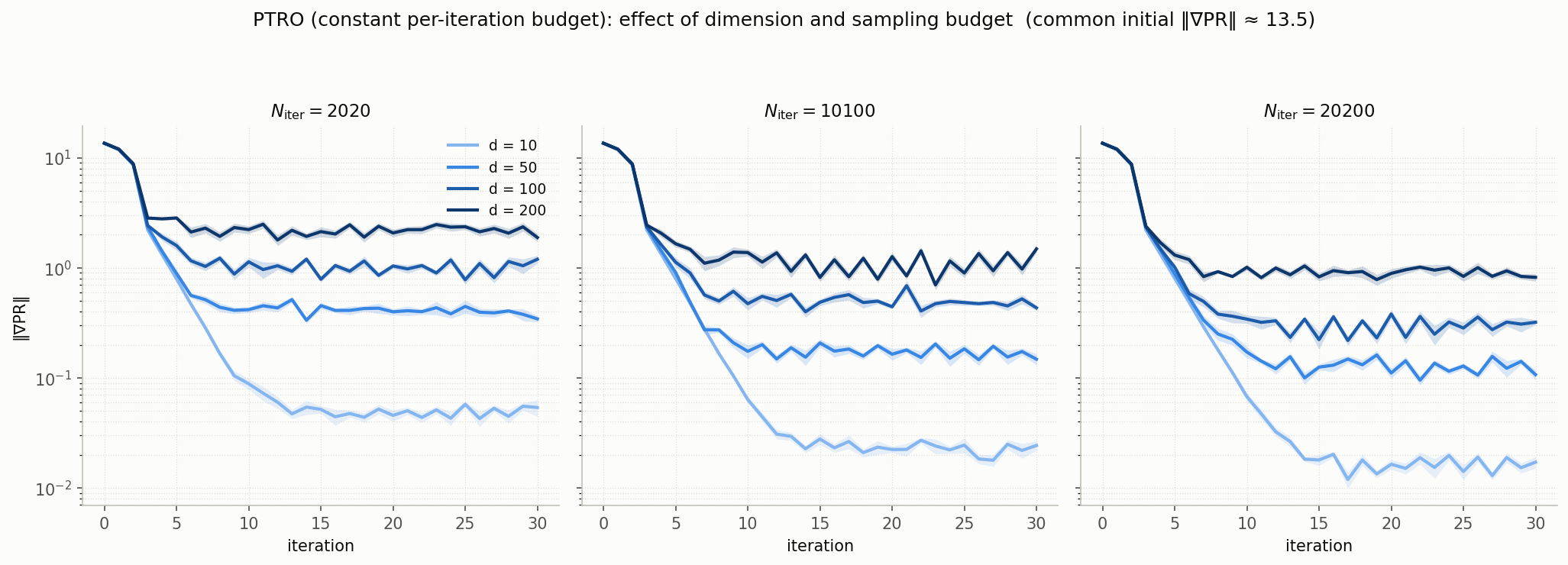}
        \caption{Constant budget.}
    \end{subfigure}

    \vspace{1mm}

    \begin{subfigure}{0.95\linewidth}
        \centering
        \includegraphics[width=\linewidth]
        {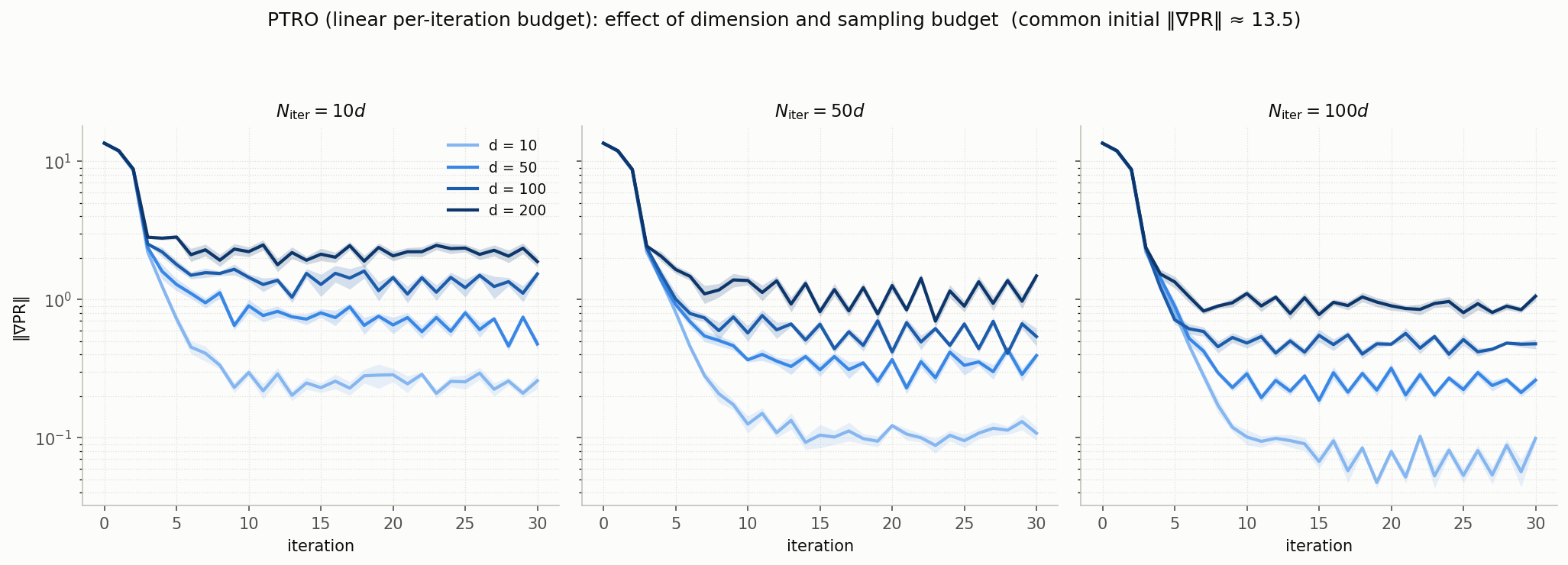}
        \caption{Sample budget scaling linearly with dimension.}
    \end{subfigure}
\begin{subfigure}{0.95\linewidth}
        \centering
        \includegraphics[width=\linewidth]
        {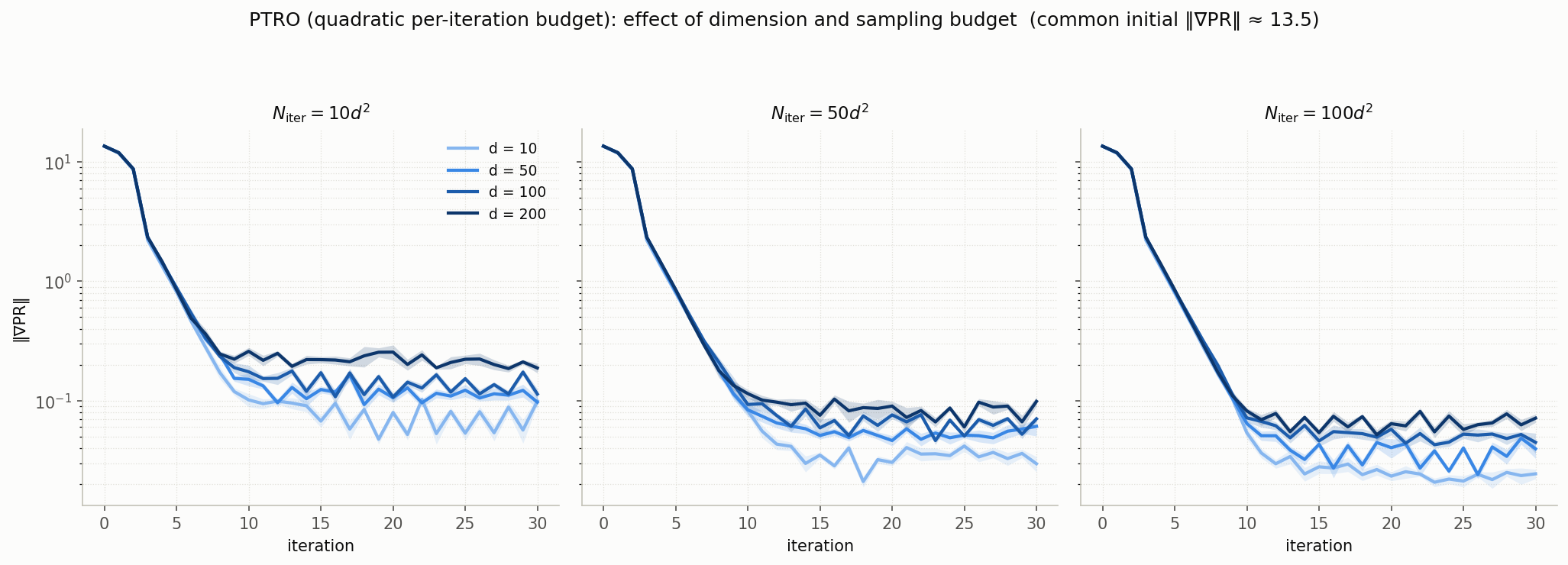}
        \caption{Sample budget scaling quadratic with dimension.}
    \end{subfigure}
    \caption{
    Effect of the sampling budget on PTRO.
    }
    \label{fig:sample_budget}
\end{figure}

\subsection{Sensitivity Experiment}
\label{appdx_subsec:sensitivity_exp}
\addcontentsline{apc}{subsection}{\protect\numberline{\thesubsection}Sensitivity Experiment}

\paragraph{Problem.}
We study the sensitivity of PTRO to
$(\gamma_{\mathrm{inc}},\gamma_{\mathrm{dec}},\eta_1,\eta_2,\alpha_0)$.

\paragraph{Setup.}
We use the same quadratic performative-risk problem as above with
$d=100$, $\sigma=4$, and $\delta_0=0$. Each function evaluation uses
$100d$ samples, and each iteration evaluates the current point, $d$
coordinate perturbations, and one trial point. We measure performance by
the number of deployments required to reach
$\|\nabla PR(x_k)\|\le 0.1$, averaged over 20 seeds.

The default parameters are
\[
(\gamma_{\mathrm{inc}},\gamma_{\mathrm{dec}},\eta_1,\eta_2,\alpha_0)
=(2.0,0.5,0.1,0.8,1.0).
\]
We conduct two studies. First, we sweep
$\gamma_{\mathrm{inc}}\in\{1.2,1.5,2.0,3.0\}$ and
$\gamma_{\mathrm{dec}}\in\{0.5,0.7,0.8,0.9\}$.
Second, fixing
$(\gamma_{\mathrm{inc}},\gamma_{\mathrm{dec}})=(2.0,0.5)$, we vary
$\eta_1$, $\eta_2$, and $\alpha_0$ one at a time.
All configurations reach the target in all 20 trials.

\paragraph{Result.}
As shown in Table~\ref{tab:sensitivity}, PTRO is relatively insensitive
to $\gamma_{\mathrm{inc}}$ and $\gamma_{\mathrm{dec}}$ over the tested
ranges. The acceptance parameters $\eta_1$ and $\eta_2$ have a more
noticeable effect: larger values reduce the number of deployments needed
to reach fine stationarity in this experiment. The initial radius
$\alpha_0$ also affects deployment efficiency. Overall, PTRO remains stable across a broad range of
trust-region parameters, and the acceptance thresholds and initial
radius mainly influence convergence speed.

\begin{table}[t]
\centering
\tiny
\setlength{\tabcolsep}{3.5pt}
\renewcommand{\arraystretch}{1.1}

\begin{tabular}{c|cccc}
\toprule
\multicolumn{5}{c}{\textbf{(a) Expansion and contraction factors}} \\
\midrule
$\gamma_{\rm dec}\backslash\gamma_{\rm inc}$
& 1.2 & 1.5 & 2.0 & 3.0 \\
\midrule
0.5 & 15.75 & 16.40 & \textbf{15.40} & 15.50 \\
0.7 & 15.85 & 16.75 & 15.50 & 16.10 \\
0.8 & 15.85 & 16.75 & 15.50 & 16.10 \\
0.9 & 15.85 & 16.75 & 15.50 & 16.10 \\
\bottomrule
\end{tabular}

\vspace{0.8em}

\begin{tabular}{c|ccccc}
\toprule
\multicolumn{6}{c}{\textbf{(b) One-at-a-time sensitivity}} \\
\midrule
Parameter
& \multicolumn{5}{c}{Value : mean deployments} \\
\midrule

$\eta_1$
& $0.05$: 15.4
& $0.1$: 15.4
& $0.25$: 15.4
& $0.5$: \textbf{10.3}
& $0.75$: 10.4
\\

$\eta_2$
& $0.1$: 15.5
& $0.4$: 15.5
& $0.8$: 15.4
& $1.2$: 13.3
& $2.0$: \textbf{12.7}
\\

$\alpha_0$
& $0.25$: 17.5
& $0.5$: 16.8
& $0.75$: 14.9
& $1.0$: 15.4
& $2.0$: \textbf{14.0}
\\

\bottomrule
\end{tabular}

\caption{
Sensitivity of PTRO to trust-region parameters at $d=100$.
Entries report the mean number of deployments required to reach
$\|\nabla PR(x)\|\le 0.1$ over 20 seeds; lower is better.
Panel (a) varies $(\gamma_{\rm inc},\gamma_{\rm dec})$ with all other
parameters fixed at their defaults. Panel (b) varies each parameter
individually with
$(\gamma_{\rm inc},\gamma_{\rm dec})=(2.0,0.5)$.
All configurations reach the target in all 20 trials.
}
\label{tab:sensitivity}
\end{table}
\clearpage

\end{document}